\documentclass{article}

\usepackage[preprint,nonatbib]{neurips_2024}

\PassOptionsToPackage{sort&compress}{natbib}

\usepackage[utf8]{inputenc} 
\usepackage[T1]{fontenc}    
\usepackage{hyperref}       
\usepackage{url}            
\usepackage{booktabs}       
\usepackage{amsfonts}      
\usepackage{nicefrac}       
\usepackage{tabularx}
\usepackage{caption}
\usepackage{makecell}
\usepackage{microtype}     
\usepackage{amsmath}
\usepackage{bbm}
\usepackage{amstext}
\usepackage{amssymb}
\usepackage{graphicx}
\usepackage{float}
\usepackage{subfigure}
\usepackage{bm}
\usepackage{multicol}
\usepackage{algorithm}
\usepackage{algorithmic}
\usepackage{wrapfig}
\usepackage{multirow}
\usepackage[table]{xcolor}
\usepackage{bigstrut}

\usepackage{setspace}
\usepackage{amsthm}
\usepackage{enumitem}
\usepackage{tablefootnote}

\title{Online DPO: Online Direct Preference Optimization with Fast-Slow Chasing}

\author{
\hspace{-32pt}
\begin{minipage}{1.1\textwidth}
\centering
  Biqing Qi\textsuperscript{1,2}, 
  Pengfei Li\textsuperscript{3}, 
  Fangyuan Li\textsuperscript{1},
  Junqi Gao\textsuperscript{3,}\thanks{Corresponding authors.},
  Kaiyan Zhang\textsuperscript{2}, 
  Bowen Zhou\textsuperscript{2,}$^*$\\
  $^1$\textnormal{Department of Control Science and Engineering, Harbin Institute of Technology,} \\
  $^2$ \textnormal{Department of Electronic Engineering, Tsinghua University,} \\
  $^3$ \textnormal{School of Mathematics, Harbin Institute of Technology} \\
  {\tt\small \{qibiqing7,lipengfei0208,jacklee19900212,gjunqi97\}@gmail.com,} \\ 
  \tt\small {zhang-ky22@mails.tsinghua.edu.cn},
  {\tt\small \{zhoubowen\}@tsinghua.edu.cn}
\end{minipage}
  }

\newtheorem{theorem}{Theorem}[subsection]
\newtheorem{lemma}{Lemma}[subsection]

\newtheorem{proposition}{Proposition}[subsection]
\newtheorem{definition}{Definition}[subsection]

\newtheorem{assumption}{Assumption}[subsection]

\begin{document}

\maketitle

\bibliographystyle{unsrt}
\begin{abstract}
Direct Preference Optimization (DPO) improves the alignment of large language models (LLMs) with human values by training directly on human preference datasets, eliminating the need for reward models. However, due to the presence of cross-domain human preferences, direct continual training can lead to catastrophic forgetting, limiting DPO's performance and efficiency. Inspired by intraspecific competition driving species evolution, we propose a \textbf{O}nline \textbf{F}ast-\textbf{S}low chasing DPO (OFS-DPO) for preference alignment, simulating competition through fast and slow chasing among models to facilitate rapid adaptation. Specifically, we first derive the regret upper bound for online learning, validating our motivation with a min-max optimization pattern. Based on this, we introduce two identical modules using Low-rank Adaptive (LoRA) with different optimization speeds to simulate intraspecific competition, and propose a new regularization term to guide their learning.
To further mitigate catastrophic forgetting in cross-domain scenarios, we extend the OFS-DPO with LoRA modules combination strategy, resulting in the \textbf{C}ross domain \textbf{O}nline \textbf{F}ast-\textbf{S}low chasing DPO (COFS-DPO). This method leverages linear combinations of fast modules parameters from different task domains, fully utilizing historical information to achive continual value alignment. Experimental results show that OFS-DPO outperforms DPO in in-domain alignment, while COFS-DPO excels in cross-domain continual learning scenarios.
\end{abstract}

\section{Introduction}
To better align Large Language Models (LLMs) with human values and prevent harmful responses, Reinforcement Learning from Human Feedback (RLHF) is commonly used in fine-tuning LLMs \cite{stiennon2020learning, christiano2017deeprlhf, bai2022training}. However, the complexity progess and the dependence on reward models in RLHF limit its practicality and efficiency \cite{bai2022training}.  To address these issues, Direct Preference Optimization (DPO) \cite{rafailov2024DPO} has emerged as an efficient alternative. The DPO employs supervised training directly based on preference data, eliminating the need for complex frameworks or specific reward models \cite{guo2024dap}, thereby simplifying model optimization for preference alignment and improving training efficiency.

However, DPO is designed for supervised training on local offline data and does not adapt well to online preference data streams \cite{guo2024dap}. Additionally, DPO cannot leverage historical information, leading to catastrophic forgetting of the original task domain during cross-domain preference alignment and resulting in overall performance degradation \cite{kirkpatrick2017overcoming, rolnick2019experience}. The cycle of forgetting and relearning significantly increases resource consumption, including computational costs and the need for re-collecting annotated data, making DPO disadvantageous in resource-constrained scenarios.

Current online learning methods for human preference alignment either involve constructing reward models, which increases resource consumption (e.g., CPPO \cite{zhang2024cppo}), or rely on feedback generated by LLMs, which lacks a flexible modular design to ensure efficient learning and memory retention \cite{guo2024dap,lee2023rlaif}. In this paper, inspired by the intraspecific competition theory \cite{bolnick2001intraspecific,thorne2003evolution}, we design competitive components with consistent optimization objectives and integrate them into the online learning to enhance the model's ability to adapt to continual changes. Our method retains the resource-efficient characteristics of DPO while improving its capability to handle continuously incoming streams of preference data.

To incorporate the concept of intraspecific competition into online preference alignment learning, we first derive the regret bounds for online learning methods \cite{agarwal2019online, hazan2020nonstochastic}. We discover that these bounds include a min-max term similar to the objective function in Generative Adversarial Networks (GANs) \cite{goodfellow2020generative}. The key difference is that, in our case, the min and max terms share the same optimization objective, closely reflecting the intraspecific competition observed in nature \cite{bolnick2001intraspecific,thorne2003evolution}, thus validating our motivation.

Furthermore, to maintain consistency with the original DPO method's adherence to human values and to prevent the policy model significantly deviating from the reference model, we retain the objective function of the original DPO. Building on this foundation, we instantiate fast and slow modules using LoRA \cite{hu2021lora} and introduce a regularization term to measure the preference probability gap between the fast and slow modules, guiding the learning of these modules. Consequently, we propose the \textbf{O}nline \textbf{F}ast-\textbf{S}low Chasing DPO (OFS-DPO) for in-domain tasks. We theoretically demonstrate that OFS-DPO achieves a lower empirical regret bound, supported by more stable gradient optimization and faster convergence.

To extend OFS-DPO to cross-domain preference alignment environments, we propose the \textbf{C}ross-domain \textbf{O}line \textbf{F}ast-\textbf{S}low Chasing DPO (COFS-DPO). Specifically, using OFS-DPO, we derive the optimal fast modules for two task domains and maintain domain-specific memories. Drawing inspiration from the human brain's capacity \cite{lake2023human,van2020brain,abs-2403-02628} for continual learning via the interplay of modular memories, our method encapsulates this mechanism through combination of LoRAs. Additionally, inspired by the conclusion of the equivalence between data shift and model parameter shift \cite{jiang2024chasing}, we theoretically derive a lower regret bound to demonstrate the effectiveness of COFS-DPO.
In experimental evaluations, our proposed OFS-DPO outperforms DPO and other competitive methods in in-domain scenarios, including controlled sentiment generation, summarization, and single-turn dialogue tasks. In cross-domain scenarios, we demonstrate that our proposed COFS-DPO significantly surpasses competitive baselines in the summarization task. In summary, our contributions are as follows: 
\begin{itemize}

    \item We propose OFS-DPO, a simple and effective method based on fast-slow LoRA modules from the novel perspective of intraspecific competition. This method introduces a regularization term to measure and guide the preference probability gap between the modules. 
   
    \item To extend OFS-DPO to cross-domain scenarios, we propose COFS-DPO, which jointly optimizes the linear combination of the optimal fast modules from different tasks. This method achieves performance comparable to theoretically optimal model parameters across the entire task domain while maintaining strong memory retention in individual domains. 
    \item We validate our proposed OFS-DPO and COFS-DPO through a theoretical analysis of the regret bounds, demonstrating their improved gradient stability and faster convergence speed.
    \vspace{-5pt}
\end{itemize}
\vspace{-10pt}
\section{Related Works}
\textbf{Direct Preference Optimization.}
The DPO aims to replace human feedback-based reinforcement learning and has found widespread application in various downstream tasks due to its resource-efficient advantages. For instance, in the multimodal domain, DPO has been beneficial for tasks such as text-to-image generation using diffusion models \cite{wallace2023diffusion}, text-to-action generation \cite{pappa2024modipo}, text-to-audio conversion \cite{majumder2024tango}, video instruction following \cite{zhang2024direct}, and translation tasks leveraging LLMs\cite{yang2023direct}.
However, DPO still has limitations that constrain its practical utility. Consequently, various improvement strategies rooted in DPO have emerged. For example, the MODPO \cite{zhou2023onepreferencefitsall} is proposed to address the requirements of multiple alignment objectives by balancing their weights. Additionally, the DPOP model \cite{pal2024smaug} introduces an enhanced DPO objective function to mitigate accuracy degradation in preference datasets with smaller edit distances, thereby improving DPO's performance on specific tasks.
Inspired by these advancements, we develop the online versions of the DPO, OFS-DPO, and COFS-DPO methods.
\textbf{Continual Learning.}
In continual learning within in-domain tasks, the need for swift adaptation to dynamic data streams has been emphasized \cite{xie2022general}. In contrast, continual learning in cross-domain scenarios faces the significant challenge of preserving previously learned task features while accommodating new ones to prevent catastrophic forgetting \cite{Simon_2022_CVPR}. Current strategies to address these challenges fall into several categories: regularization-based methods, replay-based techniques, and domain generalization methods \cite{wang2024comprehensive, Simon_2022_CVPR}.
Regularization-based methods \cite{kirkpatrick2017overcoming, aljundi2018memory, chaudhry2018riemannian, li2017learning, castro2018end} incorporate regularization terms to balance the integration of new and old knowledge, assessing the significance of various features. Replay-based methods \cite{sun2019lamol, liu2020generative,abs-2403-04140} mitigate forgetting issues in cross-domain scenarios by leveraging retained past data or experiences, thus they require resources such as memory. Additionally, ongoing researches on domain generalization \cite{pmlr-v151-rosenfeld22a, volpi2021continual, kundu2020class} aim to identify feature representations that extend beyond the training distribution while maintaining satisfactory performance on current tasks. However, these methods typically struggle when faced with substantial shifts cross-domain distributions \cite{volpi2021continual, kundu2020class}.
\vspace{-5pt}
\section{Methodology}
\subsection{Preliminaries}
In an standard online setting, we consider the distribution of data corresponding to specific human preference tasks as $\mathcal{D}$, updated within $T$ time steps. In cross-domain scenarios, we differentiate different tasks as $\mathcal{D}_1, T_1$ and $\mathcal{D}_2, T_2$. The sequence $(x_1, x_2, x_3, \ldots, x_T)$ represents samples within time $T$.
Each data point in our task setup consists of three parts: $x_i = (z, y_w, y_l)$, where $z$ is the prompt statement for the task, and $y_w$ and $y_l$ represent the desired and undesired model preferences given $z$, respectively. To ensure a fair comparison, we adhere to the settings established in \cite{zhang2024cppo}, concentrating on two domain configurations. Let $\mathcal H$ be the hypothesis class defined on distribution $\mathcal{D}$, with $h_i \in \mathcal H$ denoting a hypothesis function belonging to the hypothesis class at the $i$-th time step. The model parameter is denoted by $\theta$, with $\theta_i$ representing the model parameters at the $i$-th time step. The objective function of the DPO is denoted by $l(\theta, x)$.
\subsection{
Motivation and Theoretical Analysis}
To better understand the feasibility of the intraspecific competition motivation, we first conduct a theoretical analysis of the difference between online learning methods and the offline optimal decision by regret definition \cite{zhao2021improved}. We begin by precisely defining online expected regret to prevent any conceptual ambiguities.
\begin{definition}{(\textbf{Expected Regret})}
    \[R(T)=\mathbb{E} _{x\sim \mathcal D}{\frac{1}{T}\big[\sum_{t=1}^{T}l(h_t,x_t)- \text{min}_{h\in \mathcal H} \sum_{t=1}^{T} l(h,x_t)\big].}\]
\end{definition}

It is worth noting that $\mathcal{D}$ actually represents the distribution of task sequences in online learning, namely $\mathcal{D} = (\mathcal{D}_1, \mathcal{D}_2, \ldots, \mathcal{D}_T)$, and $x = (x_1, x_2, \ldots, x_T) \sim \mathcal{D}$.
Similar to the discussion of the regret upper bound in \cite{haghtalab2020smoothed}, we provide a similar lemma and derive the regret upper bound.
\begin{lemma}\label{boundlem}
In online learning methods, there exists a regret upper bound that includes a minimax term:
\vspace{-3pt}
\begin{equation}
    R(T)\leq O(\sqrt{\text{ln}(|\mathcal H'|)/T}) + \mathbb{E}_{x \sim \mathcal D}\frac{1}{T}\big[ \text{min}_{h'\in \mathcal H'} \text{max}_{h\in \mathcal H}(\sum_{t=1}^{T} l(h'_t,x_t)- \sum_{t=1}^{T} l(h,x_t))\big],
\end{equation}

where the first term is the regret against the best $h' \in \mathcal H'$ and $\mathcal H'$ is an infinite hypothesis class to approximate $\mathcal H$, so the second term captures how well $\mathcal H'$ approximates $\mathcal H$. 
\end{lemma}

The detailed proof of lemma \ref{boundlem} is in Appendix \ref{proofboundlem}.Let $h\in \mathcal H$ be the optimal choice in the theoretical hypothesis space. By introducing another module $\mathcal H''$ to approximate $\mathcal H$, we can derive the following inequality:
\vspace{-3pt}
\begin{equation}\label{eq2}
    \begin{split}
        R(T)\leq &O(\sqrt{\text{ln}(|\mathcal H'|)/T}) +  \mathbb{E}_{x \sim \mathcal D} 
        \frac{1}{T}\big[ \text{min}_{h'\in \mathcal H'} \text{max}_{h''\in \mathcal H''} (\sum_{t=1}^{T} l(h'_t, x_t)- \sum_{t=1}^{T} l(h'', x_t))\big]\\
        &+ \mathbb{E}_{x \sim \mathcal D} \frac{1}{T}\big[\text{min}_{h''\in \mathcal H''}\text{max}_{h\in \mathcal H}(\sum_{t=1}^{T} l(h''_t, x_t)- \sum_{t=1}^{T} l(h, x_t))\big].
    \end{split}
\end{equation}

\begin{wrapfigure}[22]{r}{0.35\textwidth}
 \vspace{1em} 
  \centering
  \includegraphics[width=0.35\textwidth]{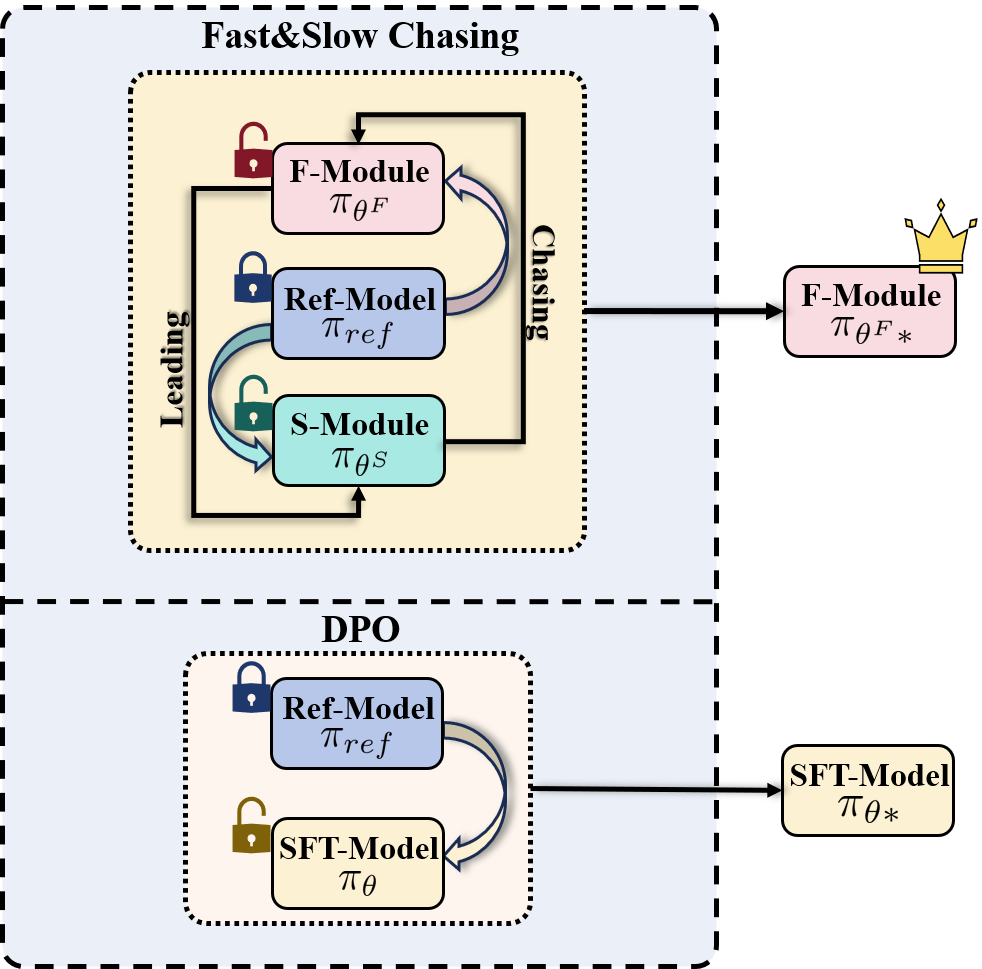}
 
  \vspace{-1.3em}
  \vspace{\baselineskip}
  \caption{\textbf{The framework of the OFS-DPO.} In the upper section, F-Module and S-Module dynamically adjust during training, while the reference model remains fixed. The lower section illustrates the framework of the original DPO.}
  \label{olfsdpopic}
\end{wrapfigure}

According to Equation \ref{eq2}, the first term is slightly affected by the learning method. Therefore, we further constrain the expected regret by minimizing the two min-max terms on the right-hand side of the inequality. \textbf{The relationship between the min term and the max term closely aligns with the competitive dynamics within natural populations, as both aim to achieve a smaller cumulative loss}.
To approximate $h'$ and $h''$ respectively, we introduce two modules for simulation: a fast module and a slow module. These modules pursue each other to approximate the best offline optimal decision $h$ \cite{zhao2021improved} ultimately optimizing the min-max term, as illustrated in Figure \ref{olfsdpopic}.
\vspace{-3pt}
\subsection{In Domain: Online Fast-Slow Chasing DPO}
\paragraph{OFS-DPO Objective Function.} 

Based on the above analysis, we understand that introducing fast and slow learning modules to simulate intraspecific competition can enhance the model's adaptability to changes, thereby improving its ability to handle continuously evolving data. However, integrating these modules renders the original objective function of the DPO \cite{rafailov2024DPO} insufficient for training these new components. To address this issue, we need to design new objective functions.
Essentially, our ultimate goal aligns with that of the DPO: to ensure the model better conforms to human value preferences while not deviating excessively from the reference model \cite{rafailov2024DPO}. Therefore, we retain the DPO objective function as a primary component of our new objective.
The DPO objective function measures the optimization gap between the learning model and the reference model. To adapt it to our new setup, we replace $\pi_{\theta}$ in the DPO objective function with $\pi_{\theta^F}$ and $\pi_{\theta^S}$ for the fast and slow modules, respectively, and denote these modules as F-module and S-module for convenience. By constructing the regularization term that measures the preference probability gap between the fast and slow modules, we introduce a chase between the modules to promote efficient optimization. Specifically, we propose the following objectives:
\begin{equation}
    \mathcal L_{DPO-new}(\theta^F)=\mathcal L_{DPO}(\theta^F)+\alpha \mathcal L_{DPO-FS},
\end{equation}
\begin{equation}
    \mathcal L_{DPO-new}(\theta^S)=\mathcal L_{DPO}(\theta^S)-\alpha \mathcal L_{DPO-FS},
\end{equation}
In the training phase, we optimize eq. (3) and (4) alternately at different frequencies. Here
\begin{equation}
    \mathcal L_{DPO}(\theta)=-\mathbb{E}_{(x,y_w,y_l) \sim \mathcal D}\left[\text{log} \sigma\left(\beta \text{log} \frac{\pi_{\theta}(y_w|x)}{\pi_{ref}(y_w|x)}-\beta \text{log} \frac{\pi_{\theta}(y_l|x)}{\pi_{ref}(y_l|x)}\right)\right],
\end{equation}
\begin{equation}
    \mathcal L_{DPO-FS}=-\mathbb E_{(x,y_w,y_l) \sim \mathcal D}\left[\text{log} \sigma\left(\beta \text{log} \frac{\pi_{\theta^F}(y_w|x)}{\pi_{\theta^S}(y_w|x)}-\beta \text{log} \frac{\pi_{\theta^F}(y_l|x)}{\pi_{\theta^S}(y_l|x)}\right)\right],
\end{equation}
where $\pi$ denotes the preference policy probability function, and $\sigma(\cdot)$ represents the logistic function. We have $\mathcal{L}_{DPO} \in (a, b) = (0, \ln 2)$. The coefficient $\alpha \in (0, 1)$ is the regularization term coefficient, and $\theta^F$ and $\theta^S$ represent the parameters of the F-module and S-module, respectively. The corresponding gradients become:
\begin{equation}\label{eq6}
    g_t^F=\nabla_{\theta^F} \mathcal L_{DPO-new}(\theta^F);\  
    g_t^S=\nabla_{\theta^S} \mathcal L_{DPO-new}(\theta^S).
\end{equation}
\paragraph{OFS-DPO.} 
Based on the above analysis, we proceed to formally construct OFS-DPO. Specifically, incorporating diverse LoRA modules into the reference model, we initialize F-module and S-module respectively. Using the update gradients mentioned in eq. (7), we update the parameters $\theta^F$ and $\theta^S$, respectively.
Every $k$ time steps, we swap the roles of the F-module and the S-module if $\mathcal{L}_{\text{DPO}}(\theta^F) > \mathcal{L}_{\text{DPO}}(\theta^S)$, to ensure that the module with the best performance is designated as the F-module. Otherwise, we maintain the current setup and proceed to the next update cycle, thereby achieving the fast-slow chasing effect during training. The method is summarized as Algorithm \ref{algo1} in Appendix \ref{proposed algorithms}.
For the instantiation of the method, we require an appropriate tool to create fast-slow modules with different optimization speeds, allowing for flexible dynamic switching during training. LoRA \cite{hu2021lora}, currently the most commonly used efficient fine-tuning strategy for LLMs, perfectly meets our requirements and mitigates the cost increase associated with introducing two modules.
\paragraph{Theoretical analysis of OFS-DPO.}
To further validate the theoretical effectiveness of our proposed OFS-DPO, we employ regret analysis to demonstrate that our method can achieve a lower empirical regret bound more rapidly in in-domain tasks. We first present the empirical distribution of the regret.
\begin{definition}{(\textbf{Experience of Regret})}
    \begin{equation}
        R(T)=\frac{1}{T}\sum_{i=1}^{T} [l(\theta_T,x_i)- l(\theta^*,x_i)],
    \end{equation}
    where $\theta^* \triangleq \text{argmin}_\theta \frac{1}{T}\sum_{i=1}^{T}l(\theta,x_i) $, $x_i \sim \mathcal D$, $i\in \{1,2,\cdots,T\}$ and $\mathcal D$ is the data distribution.
\end{definition}
In our method, at regular update intervals, we compare $\mathcal{L}_{DPO}(\theta^F)$ and $\mathcal{L}_{DPO}(\theta^S)$ to ensure that the better-performing model is designated as the F-module. For convenience, let $\theta$ and $w$ represent the parameters of the F-module and S-module, respectively. Consequently, the optimal module parameters at each time step should satisfy the following condition:
\begin{equation}\label{eq8}
\hat\theta_i=\text{argmin}_\theta(l(\theta_i,x_i),l(w_i,x_i)).
\end{equation}
Continuing, we can express the empirical regret of the OFS-DPO in the following form:
\begin{equation}\label{eq9}
    \hat R(T)=\sum_{i=1}^{T} \frac{1}{T}[l(\hat \theta_T,x_i)- l(\theta^*,x_i)].
\end{equation}
Before conducting a detailed quantitative analysis of the empirical regret \ref{eq9}, we give some boundedness assumptions for the gradient at current step and task-specific module parameters.
\begin{assumption} \label{gb}
    \textbf{(Gradient boundedness)} Denote $g_i = \nabla_\theta l(\theta_i,x_i)$, $i=1,2,\cdots,T$, then $||g_i||_2\le G$, where G is a positive constant. 
\end{assumption}
\begin{assumption} \label{pb}
	\textbf{(Model parameters boundedness)} Suppose $||\theta_n-\theta_m||_2 \le d$,  $||w_n-w_m||_2 \le d, \forall n,m \in (1,...T)$, where $d$ is a positive constant.
\end{assumption}
Under assumptions \ref{gb} and \ref{pb}, we can derive the following theorem.
\begin{theorem}\label{singletaskbound}
    Within proposed OFS-DPO, a lower empirical regret bound can be attained, with a probability $1-\delta$, where $\delta = 2(T-1)\delta_0- (T-1)(2T-3)(\delta_0)^2[1-\delta_0]^{2T-4}$,
    and $\delta_0\in \ (0,1)$.
    \begin{equation}\label{eq11}
	R(T) \ge l(\theta_1,x_1)-\frac{1}{T}\sum_{i=1}^{T}l(\theta^*,x_i)-\big[2-\frac{1}{T}+(1-\frac{1}{T})\mathbbm{1}_{\{mode=FS\}}\big]Gd-2(1-\frac{1}{T})\text{ln}2\sqrt{-\frac{\text{ln}\delta_0}{2}},
    \end{equation}
    where $\mathbbm{1}_{\{\text{mode}=FS\}}$ represents whether to introduce fast and slow modules.  
\end{theorem}
From Theorem \ref{singletaskbound}, we observe that with the introduction of the fast-slow mode, i.e., $\mathbbm{1}_{\{\text{mode}=FS\}}=1$, the right-hand side of Inequality \ref{eq11} decreases further by $(1-\frac{1}{T})Gd$. This indicates that our proposed OFS-DPO achieves a lower bound on empirical regret. More detailed proofs can be found in Appendix \ref{proofofolfs}.

\paragraph{More Stable Gradient.}
Building upon a superior lower bound on empirical regret, we further demonstrate through a proposition that incorporating the $L_{DPO-FS}$ regularization term results in more stable gradient information.
\begin{proposition}\label{stablegradients}
    As training progresses, $\forall \epsilon>0$ such that $\nabla_\theta \mathcal{L}_{DPO}(\theta) < \epsilon$, while $\nabla_{\theta^F} \mathcal{L}_{DPO-new}(\theta^F) > \epsilon$. In other words, the original DPO experiences significantly diminish gradients as training continues, leading to a lack of update momentum. Introducing the $\mathcal{L}_{DPO-FS}$ regularization term can address this issue.
\end{proposition}

The above proposition explains, from the perspective of the model update mechanism, why OFS-DPO can achieve better performance than the original DPO. \textbf{A key reason is that our method maintains more sustained gradient update momentum}.
\subsection{Cross Domain: Online Fast Slow Chasing DPO}
\begin{figure}[t]
  \centering
  \includegraphics[width=0.85\textwidth]{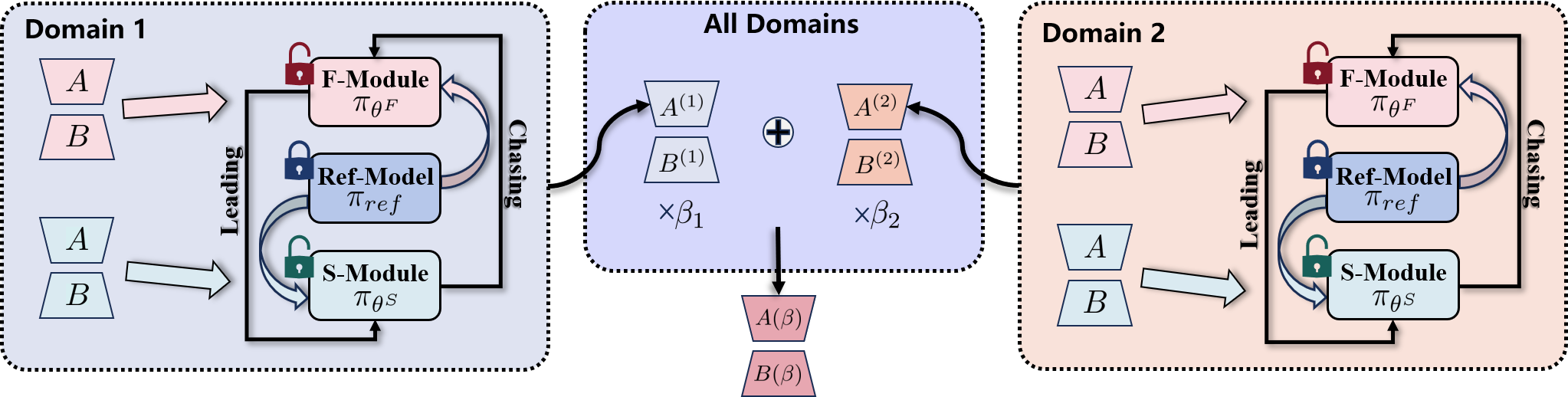}
  \vspace{\baselineskip}
  \caption{\textbf{The framework of the COFS-DPO.} Instantiate the fast-slow modules with LoRAs separately in different task domains to obtain the optimal LoRA module in each domain. Subsequently, we seek the optimal linear combination $(\beta_1,\beta_2)$ across all task domains.}
  \label{csfsdpopic}
  \vspace{-10pt}
\end{figure}

\paragraph{COFS-DPO.}
Furthermore, we extend OFS-DPO to cross-domain scenarios. The main distinction between the COFS-DPO and the in-domain setting lies in balancing the importance of information obtained from different task domains. This necessitates preserving and integrating historical data from various domains in a specific manner to mitigate catastrophic forgetting in cross-domain scenarios. Inspired by the human brain's ability \cite{lake2023human,van2020brain,abs-2403-02628} to achieve continual learning through the interaction of modular memories, we model this process using the combination of LoRAs. Thus, we achieve the COFS-DPO method to retain crucial historical information across tasks, as illustrated in Figure \ref{csfsdpopic}.

Specifically, in the cross domain scenario, we first use the OFS-DPO to obtain the final F-modules for two task domains, retaining a random subset of domain-specific memories $M_1$ and $M_2$. Subsequently, we compute the optimal combination of F-modules over the joint memory distribution $(M_1, M_2)$ to achieve the best performance. The detailed procedure is outlined in Algorithm \ref{algo2} in the Appendix \ref{proposed algorithms}.
\paragraph{Theoretical analysis of COFS-DPO.}
Next, we provide a theoretical analysis to demonstrate the effectiveness of the proposed COFS-DPO. To ensure clarity, we first standardize the use of symbols: let $s_i^{(k)}$ represent the sample at the $i$-th moment from distribution $\mathcal{D}_k$, where $k=1,2$ and $i=0,1,2,\dots,T_k$. Let $\theta_i$ represent the model parameters at moment $i$. Specifically, $\theta^*$ denotes the optimal parameters for the overall task distribution, $\theta^{(1)}$ denotes the optimal parameters for distribution $\mathcal{D}_1$, and $\theta^{(2)}$ denotes the optimal parameters for distribution $\mathcal{D}_2$. Consider describing the relationship between these parameters in an incremental manner: 
$\theta^*=\theta_0+\Delta\theta^*,\ \theta^{(1)}=\theta_0+\Delta\theta^{(1)},\ \theta^{(2)}=\theta_0+\Delta\theta^{(2)}$.

Building upon these symbol definitions and the previously established lower bound on single-task regret (Equation \ref{singletaskbound}), we can derive Theorem \ref{multitaskbound} for dual-task scenarios. This foundation also enables the extension to cross domain scenarios.
\begin{definition}\label{dualtaskregret}
    \textbf{(Experience regret of cross-domain tasks: Dual-Task Regret)}
    \begin{equation}
        R(T_1,T_2) = \frac{1}{T_1}\sum_{i=1}^{T_1}l(\theta_{T_1},s_{i}^{(1)})+\frac{1}{T_2}\sum_{j=1}^{T_2}l(\theta_{T_2},s_{j}^{(2)})-\frac{1}{T_1}\sum_{i=1}^{T_1}l(\theta^*,s_{i}^{(1)})-\frac{1}{T_2}\sum_{j=1}^{T_2}l(\theta^*,s_{j}^{(2)}).
    \end{equation}
\end{definition}
Given the definition of dual-task regret, we derive a lower regret bound similar to the in-domain case.
\begin{theorem}\label{multitaskbound}   
    Under the settings of this paper, the dual-task regret also has a lower empirical regret bound.
    \begin{equation}\label{eq13}
        \begin{split}
            R(T_1,T_2) \ge\  &l_1(s_1^{(1)},s_1^{(2)})-B(T_1,T_2)-2(1-\frac{T_1+T_2}{T_1T_2})c\\
            &-[6-\frac{T_1+T_2}{T_1T_2}+(2-\frac{T_1+T_2}{T_1T_2})\mathbbm{1}_{\{mode=FS\}}]Gd,
        \end{split}
    \end{equation}
    where $B(T_1,T_2)=\frac{1}{T_1}\sum_{i=1}^{T_1}l(\theta^{(1)},s_{i}^{(1)})+\frac{1}{T_2}\sum_{j=1}^{T_2}l(\theta^{(2)},s_{j}^{(2)})$, $  
l_1(s_1^{(1)},s_1^{(2)})=l(\theta_1,s_1^{(1)})+l(\theta_2,s_1^{(2)})$,
    $c=\max\{\text{ln}2\sqrt{-\frac{\text{ln}\delta_1}{2}},\text{ln}2\sqrt{-\frac{\text{ln}\delta_2}{2}}\},\ where\ \delta_1,\delta_2\in (0,1)$, $\mathbbm{1}_{\{mode=FS\}}$ represents whether to introduce fast- slow modules.
\end{theorem}
Theorem \ref{multitaskbound} shows that with the introduction of the fast-slow mode, the right side of Inequality \ref{eq13} is further reduced by $(2 - \frac{T_1 + T_2}{T_1 T_2})Gd$. This indicates that COFS-DPO, like OFS-DPO, also achieves a lower empirical regret bound.
\section{Experiments}
\vspace{-10pt}
\begin{table}[h]
    \caption{\textbf{GPT-4 win rates for three in-domain tasks.} We set the rank of LoRA between 16 and 256 to observe its impact on the model fine-tuning results. We report results over 5 trials. More implementation details are available in the Appendix \ref{Experimental Details}.}
    \label{indolab}
     \vspace{\baselineskip}
    \renewcommand{\arraystretch}{0.3}
    \centering
    \resizebox{0.95\textwidth}{!}{
    \begin{tabular}{ccccc}%
    \toprule%
    \textbf{LoRA rank} & \textbf{Method} & \textbf{Controlled Sentiment Generation} & \textbf{Summarization} & \textbf{Single-turn Dialogue} \\
    \midrule
    \multirow{4}{*}{16}& SFT & 0.307±0.042 & 0.128±0.036 & 0.132±0.033 \\
    \cmidrule[0.5pt]{2-5}
     & DPO &0.488±0.037 & 0.194±0.014& 0.192±0.012 \\
     & PPO & 0.358±0.056 & 0.136±0.048 & 0.147±0.052 \\
     & OFS-DPO (Ours) & \textbf{0.568±0.034}& \textbf{0.223±0.033} & \textbf{0.231±0.014} \\
    \midrule
    \multirow{4}{*}{256} & SFT & 0.313±0.035 & 0.138±0.036 & 0.138±0.032 \\
    \cmidrule[0.5pt]{2-5}
     & DPO &0.494±0.032 &0.211±0.028 & 0.203±0.024\\
     & PPO & 0.364±0.039 & 0.148±0.044 & 0.155±0.042 \\
     & OFS-DPO (Ours) & \textbf{0.571±0.043} & \textbf{0.234±0.021} &\textbf{0.253±0.017}  \\
    \bottomrule
    \end{tabular}}
    \vspace{-3pt}
\end{table}
\vspace{-3pt}
\subsection{Experimental settings}
The experiments are primarily divided into two parts. The first part aims to validate the effectiveness of the OFS-DPO in adapting to online data streams of the in-domain tasks. The second part focuses on evaluating the COFS-DPO's capability to retain model memory in cross-domain scenarios.

\textbf{In-Domain Task Setting.}
In the online in-domain preference alignment, we maintain consistent experimental settings with DPO \cite{rafailov2024DPO}, validating our method's effectiveness across three tasks: controlled sentiment generation, summarization, and single-turn dialogue.
For the controlled sentiment generation task, we use the IMDB dataset. Initially, we fine-tune the GPT-2 LLM \cite{radford2019language} on the IMDB training dataset to obtain the Supervised Fine-Tuning (SFT) \cite{ziegler2019fine} model. Subsequently, using positive and negative reviews from the dataset as preference data, we further fine-tune the SFT model using the OFS-DPO to align it with human value preferences.
In the summarization task, we use the TL;DR Summarize dataset \cite{stiennon2020learning} with human preferences. We fine-tune the GPT-J \cite{gpt-j} on the summarization dataset as the SFT model and then conduct online value preference alignment using our method.
For single-turn dialogue, we fine-tune the Pythia-2.8B model\footnote{https://huggingface.co/EleutherAI/pythia-2.8b} on the Anthropic Helpful and Harmless dialogue dataset\footnote{https://huggingface.co/datasets/Anthropic/hh-rlhf\label{hh}} as the SFT model. Subsequently, we train the SFT model with the OFS-DPO.

In the evaluation process, we introduce PPO and DPO, along with the corresponding SFT models for each task, as baseline models to compare against our method under identical training settings. Specifically, alignment with the previous studies\cite{rafailov2024DPO,guo2024dap,zhang2024cppo,chen2023exploring}, we use GPT-4 as a surrogate human evaluator to assess the quality of model-generated content and compare it with preferences extracted from authentic datasets. The resulting win rate serves as a metric to quantify the effectiveness of model alignment. More detailed experimental configurations are provided in the appendix \ref{Experimental Details}.

\textbf{Cross-Domain Task Setting.} Based on the experimental setup in CPPO \cite{zhang2024cppo}, we design our cross-domain experiment named "Summary." We utilize two datasets: the Reddit TL;DR dataset for SFT and the human preference dataset \footnote{https://huggingface.co/datasets/CarperAI/openai\_summarize\_comparisons} provided by CarperAI for RLHF. To validate our method's continual learning capability, we partition these datasets into two domains based on post types: "relationships" and "others." We denote the "relationships" domain as Task-1, comprising a single category, while the "others" domain is denoted as Task-2, consisting of the remaining 28 categories.
To ensure a fair comparison with CPPO, we adopt the same experimental settings. We train a GPT-2 Small (GPT2-s) model \footnote{https://huggingface.co/openai-community/gpt2} with 124M parameters using Task-1 data from the Reddit TL;DR dataset for 5 epochs as the SFT model. Additionally, we train the LLaMA3 (8B) model \cite{llama3modelcard} for further testing. We fine-tune the SFT model combined with LoRA on the Task-1 data of the human preference dataset to obtain $\theta^{(1)}$. Then, we fine-tune it on the Task-2 data after initializing the model to obtain $\theta^{(2)}$. We retain a small amount of data from different tasks to provide COFS-DPO for the combined optimization of $\Delta \theta^{(1)}$ and $\Delta \theta^{(2)}$.

To evaluate model alignment performance, 
{alignment with the previous works \cite{zhang2024cppo,gao2023scaling}, we fine-tune the GPT-J (6.7B) model on the entire human preferences dataset as a reference preference model (rPM). We use ROUGE \cite{lin2004rouge} and rPM scores (rPMS)  \cite{zhang2024cppo} to measure the model's alignment with current data and use SFR metric \cite{zhang2024cppo} to measure forgetting rate of old data. Table \ref{metrics} in Appendix \ref{Experimental Details} presents the evaluation metrics for each task.

\begin{table}[t]

    \caption{\textbf{The main evaluation results of cross-domain tasks are presented.} Hyperparameters for CPPO can be defined in two ways: heuristic or learnable. For our comparison with COFS-DPO, we use heuristic CPPO (CPPOH). Experiments were conducted using GPT2-s and  LLaMA3. The rank of LoRA was set to 16 and 256. We report results over 5 trials. More implementation details are available in the Appendix \ref{Experimental Details}.}
    \label{crdolab}
     \vspace{\baselineskip}
    \centering
    \renewcommand{\arraystretch}{0.7}
    \resizebox{0.9\linewidth}{!}{
    \begin{tabular}{cccccccccc}
    \toprule%
    \multirow{2}{*}{\textbf{Model}} & \multirow{2}{*}{\textbf{LoRA rank}} & \multirow{2}{*}{\textbf{Method}} & \multicolumn{2}{c}{\textbf{Task-1}} &  \multicolumn{5}{c}{\textbf{Final}}\\
    \cmidrule(r){4-5} \cmidrule(r){6-10}
    & & & \textbf{rPMS$_1$}($\uparrow$) & \textbf{Rouge$_1$}($\uparrow$)& \textbf{rPMS$_1$}($\uparrow$)& \textbf{Rouge$_1$}($\uparrow$)& \textbf{SFR}($\downarrow$) & \textbf{rPMS$_2$}($\uparrow$) & \textbf{Rouge$_2$}($\uparrow$) \\
    \midrule
    \multirow{22}{*}{GPT2-s}& \multirow{10}{*}{16} & DPO \cite{rafailov2024DPO} & 5.750±0.124 & 0.228±0.009 & 5.773±0.132 & 0.230±0.008 & -0.023±0.012 & 4.785±0.121 & 0.167±0.019 \\
    &  & PPO+EWC \cite{kirkpatrick2017overcoming} & 4.932±0.117 & 0.203±0.011 & 4.983±0.108 & 0.207±0.014 & -0.051±0.014 & 4.505±0.118 & 0.137±0.014 \\
    &  & PPO+LwF \cite{li2017learning} & 4.890±0.122 & 0.199±0.021 & 4.953±0.118 & 0.202±0.010 & -0.063±0.004 & 4.533±0.114 & 0.128±0.008 \\
    &  & PPO+TFCL \cite{Aljundi_2019_CVPR} & 4.934±0.148 & 0.217±0.018 & 4.988±0.220 & 0.217±0.013 & -0.054±0.010 & 4.524±0.113 & 0.135±0.012 \\
    & & PC\cite{kaplanis2019policy} & 4.811±0.210 & 0.204±0.031 & 4.845±0.164 & 0.216±0.008 & -0.034±0.013 & 4.574±0.120 & 0.149±0.011 \\ 
    &  & HN-PPO \cite{schopf2022hypernetwork} & 4.945±0.151 & 0.218±0.013 & 4.992±0.211 & 0.204±0.011 & -0.047±0.012 & 4.531±0.147 & 0.136±0.012 \\
    &  & NLPO \cite{Ramamurthy2022IsRL} & 4.931±0.121 & 0.203±0.022 & 4.987±0.136 & 0.208±0.033 & -0.056±0.007 & 4.482±0.124 & 0.136±0.019\\
    & & CPPO-H \cite{zhang2024cppo} & 5.059±0.212 & 0.211±0.012 & 5.410±0.189 & 0.213±0.010 & -0.351±0.015 & 4.629±0.127 & 0.165±0.017 \\
    &  & COFS-DPO (Ours) & \textbf{5.756±0.212} & \textbf{0.230±0.010} & \textbf{6.398±0.224} & \textbf{0.234±0.007} & \textbf{-0.642±0.014} & \textbf{5.641±0.151} & \textbf{0.174±0.021} \\
    \cmidrule(r){2-10}
     & \multirow{10}{*}{256} & DPO \cite{rafailov2024DPO} & 5.754±0.183 & 0.230±0.010 & 5.766±0.215 & 0.230±0.013 & -0.012±0.009 & 4.793±0.131 & 0.168±0.019 \\
      &  & PPO+EWC \cite{kirkpatrick2017overcoming} & 4.944±0.161 & 0.208±0.011 & 4.997±0.172 & 0.211±0.015 & -0.053±0.007 & 4.505±0.249 & 0.137±0.010 \\
    &  & PPO+LwF \cite{li2017learning} & 4.907±0.198 & 0.203±0.015 & 4.969±0.167 & 0.210±0.013 & -0.062±0.007 & 4.562±0.154 & 0.151±0.007 \\
     &  & PPO+TFCL \cite{Aljundi_2019_CVPR} & 5.068±0.185 & 0.213±0.012 & 5.142±0.231 & 0.204±0.013 & -0.074±0.011 & 4.563±0.214 & 0.145±0.014 \\
    & & PC\cite{kaplanis2019policy} & 4.923±0.245 & 0.207±0.023 & 4.981±0.178 & 0.237±0.016 & -0.058±0.011 & 4.558±0.214 & 0.119±0.008 \\
    &  & HN-PPO \cite{schopf2022hypernetwork} & 5.072±0.235 & 0.214±0.011 & 5.153±0.234 & 0.208±0.019 & -0.081±0.009 & 4.575±0.212 & 0.157±0.012 \\
    &  & NLPO \cite{Ramamurthy2022IsRL} & 5.113±0.261 & 0.209±0.013 & 5.201±0.246 & 0.206±0.020 & -0.088±0.011 & 4.501±0.196 & 0.146±0.014 \\
    &  & CPPO-H \cite{zhang2024cppo} & 5.270±0.284 & 0.211±0.016 & 5.618±0.258 & 0.217±0.021 & -0.348±0.014 & 4.664±0.188 & 0.134±0.016 \\ 
    &  & COFS-DPO (Ours) & \textbf{5.816±0.248} & \textbf{0.234±0.012} & \textbf{6.430±0.245} & \textbf{0.241±0.013} & \textbf{-0.614±0.023} & \textbf{5.664±0.213} & \textbf{0.178±0.017} \\
    \midrule
    \multirow{22}{*}{Llama3}& \multirow{10}{*}{16} & DPO \cite{rafailov2024DPO} & 5.809±0.215 & 0.110±0.013 & 5.816±0.236 & 0.113±0.015 & -0.007±0.014 & 5.473±0.267 & 0.099±0.016 \\
    &  & PPO+EWC \cite{kirkpatrick2017overcoming} & 5.076±0.221 & 0.102±0.016 & 5.160±0.186 & 0.113±0.027 & -0.084±0.012 & 5.050±0.271 & 0.102±0.011 \\
    &  & PPO+LwF \cite{li2017learning} & 5.007±0.232 & 0.103±0.012 & 5.109±0.245 & 0.110±0.012 & -0.102±0.016 & 4.601±0.169 & 0.111±0.014 \\
     &  & PPO+TFCL \cite{Aljundi_2019_CVPR} & 5.082±0.231 & 0.108±0.017 & 5.171±0.235 & 0.113±0.009 & -0.089±0.009 & 4.624±0.123 & 0.112±0.011 \\
     & & PC\cite{kaplanis2019policy} & 5.012±0.271 & 0.094±0.016 & 5.104±0.244 & 0.112±0.013 & -0.092±0.010 & 4.573±0.239 & 0.082±0.014 \\
    &  & HN-PPO \cite{schopf2022hypernetwork} & 5.098±0.258 & 0.098±0.014 & 5.168±0.198 & 0.106±0.019 & -0.070±0.015 & 4.627±0.211 & 0.127±0.010 \\
    &  & NLPO \cite{Ramamurthy2022IsRL} & 5.018±0.219 & 0.087±0.013 & 5.101±0.272 & 0.108±0.012 & -0.083±0.008 & 4.594±0.222 & 0.110±0.016 \\
    &  & CPPO-H \cite{zhang2024cppo} & 5.121±0.214 & 0.096±0.011 & 5.449±0.261 & 0.101±0.009 & -0.328±0.021 & 5.318±0.264 & 0.089±0.013 \\ 
    &  &  COFS-DPO (Ours) & \textbf{5.867±0.220} & \textbf{0.115±0.014} & \textbf{7.285±0.288} & \textbf{0.159±0.012} & \textbf{-1.418±0.026} & \textbf{7.094±0.301} & \textbf{0.139±0.010} \\
     \cmidrule(r){2-10}
     & \multirow{10}{*}{256} & DPO \cite{rafailov2024DPO} & 5.801±0.271 & 0.114±0.023 & 5.805±0.243 & 0.114±0.012 & -0.004±0.007 & 5.477±0.264 & 0.099±0.010 \\
     &  & PPO+EWC \cite{kirkpatrick2017overcoming} & 5.121±0.215 & 0.101±0.016 & 5.220±0.275 & 0.105±0.014 & -0.099±0.017 & 5.216±0.248 & 0.115±0.010 \\
      &  & PPO+LwF \cite{li2017learning} & 5.107±0.214 & 0.098±0.013 & 5.201±0.237 & 0.110±0.010 & -0.094±0.014 & 5.203±0.235 & 0.108±0.011 \\
     &  & PPO+TFCL \cite{Aljundi_2019_CVPR} & 5.172±0.233 & 0.109±0.012 & 5.263±0.269 & 0.116±0.035 & -0.091±0.015 & 5.278±0.249 & 0.094±0.031 \\
     & & PC\cite{kaplanis2019policy} & 4.893±0.198 & 0.101±0.023 & 4.980±0.251 & 0.107±0.012 & -0.087±0.022 & 4.995±0.276 & 0.056±0.005 \\ 
    &  & HN-PPO \cite{schopf2022hypernetwork} & 5.168±0.314 & 0.111±0.018 & 5.235±0.341 & 0.109±0.014 & -0.067±0.022 & 5.280±0.361 & 0.096±0.021 \\
    &  & NLPO \cite{Ramamurthy2022IsRL} & 5.096±0.277 & 0.092±0.019 & 5.167±0.301 & 0.108±0.024 & -0.071±0.014 & 5.236±0.267 & 0.038±0.012 \\
    &  & CPPO-H \cite{zhang2024cppo} & 5.322±0.255 & 0.102±0.011 & 5.657±0.248 & 0.097±0.014 & -0.335±0.022 & 5.351±0.257 & 0.060±0.007 \\
    &  &  COFS-DPO (Ours) & \textbf{5.895±0.269} & \textbf{0.116±0.011} & \textbf{7.508±0.285} & \textbf{0.163±0.014} & \textbf{-1.613±0.022} & \textbf{7.317±0.275} & \textbf{0.143±0.014} \\
    \bottomrule
    \end{tabular}}

\vspace{-3pt}
\end{table}
\vspace{-5pt}
\subsection{Evaluation results on in-domain tasks}

Table \ref{indolab} presents the performance of the OFS-DPO compared to traditional DPO, PPO, and SFT models under an online in-domain task data stream. Using GPT-4 as a human proxy \cite{rafailov2024DPO,guo2024dap,zhang2024cppo,chen2023exploring}, we evaluate the model-generated content against actual preference data.
The results indicate that the OFS-DPO consistently outperforms the DPO and PPO. Specifically, in the controlled emotion generation task, OFS-DPO achieves approximately an 8\% improvement in win rate across different LoRA ranks. In the single-turn dialogue task, it demonstrates an improvement of approximately 5\% in win rates. This shows the superior alignment effectiveness of the OFS-DPO in various tasks. 

\subsection{Evaluation results on cross-domain tasks}

Table \ref{crdolab} presents the results of continual learning for the summarization task based on human preference datasets. In this experiment, we used GPT2-s and LLaMA3 as our fundamental models. After training on Task 1, we evaluated the model's performance on Task 1 using rPMS and Rouge metrics. We then continue training on Task 2 and re-evaluate the model's performance on both Task 1 and Task 2. The ability of the model to overcome catastrophic forgetting was assessed by examining the changes in rPMS for Task 1 before and after training on Task 2.
The results indicate that when using GPT2-s as the base model, the COFS-DPO achieves an SFR metric of around -0.6, which significantly surpasses the memory retention performance of all baselines. When using Llama3 as the base model, the corresponding SFR metric is around -1.5, nearly twice as good as the best-performing PPO variant. This shows the superior memory retention capabilities of our method.

\subsection{Ablation Studies}

To validate the impact of the coefficient $\alpha$ in the regularization term of our objective function on the win rate of models in the controlled sentiment generation task, we designed experiments with $\alpha$ ranging from 0 to 0.9. The results, illustrated in the left panel of Figure \ref{ablapic}, indicate that the model's win rate remains stable around 50\%, even in the least favorable scenario ($\alpha = 0$), demonstrating the stability of our method.
The second panel from the left in Figure \ref{ablapic} investigates the effect of varying the learning rate multiples between fast-slow modules on training effectiveness. Across various LoRA rank settings, OFS-DPO significantly outperform the baseline PPO, maintaining a lead of at least 10\% even in the worst-case scenario. This suggests that our method is robust across different learning rate configurations.
As illustrated in the second panel from the right in Figure \ref{ablapic}, increasing the batch size and the contrastive update period between fast-slow modules leads to a more stable win rate for our models. This demonstrates the positive impact of these adjustments on model performance.
The rightmost panel in Figure \ref{ablapic} shows that the gradient norms of the OFS-DPO exhibit more sustained stability compared to those of the original DPO loss. This provides experimental evidence of our method's superiority in maintaining gradient stability, further supporting its overall effectiveness.

\begin{figure}[t]
\centering
    \includegraphics[width=0.9\textwidth]{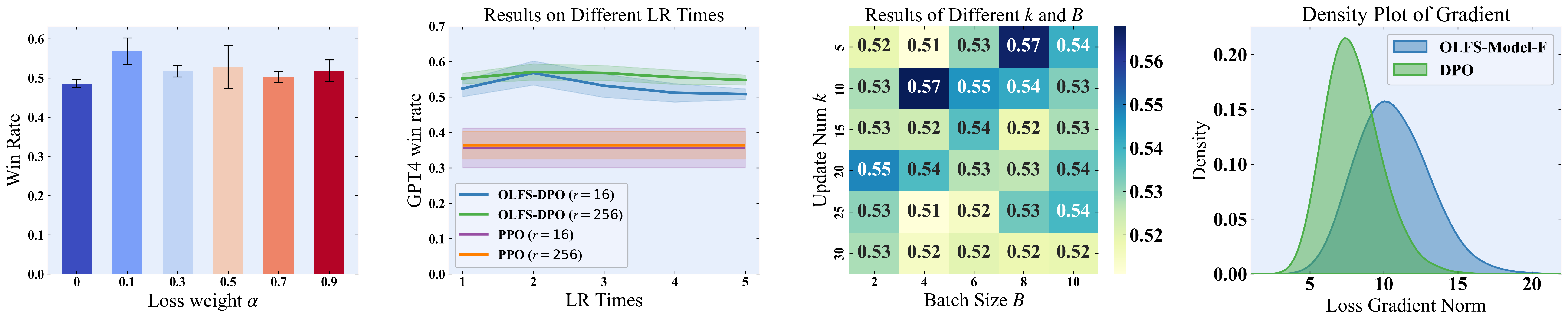} 
    
    \caption{\textbf{All ablation results are based on IMDB.} From left to right: Win rates with different choices of the regularization coefficient $\alpha$; win rates comparing OFS-DPO and PPO under varying learning rate multipliers between fast-slow modules; the influence of batch size and the contrast period \( k \) between fast-slow modules on win rates; and kernel density estimates of the loss gradients from the original DPO and the OFS-DPO during the training process.}
    \label{ablapic}
    \vspace{-15pt}
\end{figure}

\begin{wraptable}[10]{r}{0.5\textwidth}
\vspace{-15pt}
\centering
\caption{Results of COFS-DPO with different sample sizes.}
\label{tab:my_label}
\small
\begin{tabular}{cccc} 
\toprule
\textbf{Sample Num} & $\mathbf{rPMS_1}$ & $\mathbf{SFR}$ & $\mathbf{rPMS_2}$ \\
\midrule
100 & 5.988 & -0.242 & 5.156 \\
200 & 6.180 & -0.434 & 5.340 \\
500 & 6.398 & -0.652 & 5.641 \\
1000 & 6.367 & -0.621 & 5.590 \\
2000 & 6.312 & -0.641 & 5.605 \\
\bottomrule
\end{tabular}
\end{wraptable}

To investigate the impact of retaining specific domain samples on the memory retention capability of models in cross-domain tasks, we designed validation experiments with total sample sizes of 100, 200, 500, and 1000 for two tasks, as shown in Table \ref{tab:my_label}. The results indicate that when the total sample size increases to 500, the COFS-DPO achieves optimal retention of historical preference information. Beyond this sample size, further increases do not enhance the method's performance. This finding suggests that within our proposed framework, it is unnecessary to retain a large number of specific domain samples to achieve excellent results.
\vspace{-5pt}
\section{Conclusion}

In this work, inspired by intraspecific competition theory, we propose a simple and effective OFS-DPO, which leverages the competition between fast and slow modules under the same objective to achieve continual preference learning, with theoretical guarantees in terms of regret bounds and gradient stability. Furthermore, to extend OFS-DPO to cross domain settings, we introduce COFS-DPO, which achieves preference alignment by replacing the optimal LoRA for a task domain with a linear combination of LoRAs from different domains, validated both experimentally and theoretically.
Our work demonstrates that proposed methods based on intraspecific competition provide new insights and solutions for online human preference alignment tasks and have the potential for broad applicability across multi domains. 

\clearpage
\pagebreak
\bibliography{nips}

\begin{thebibliography}{10}

\bibitem{stiennon2020learning}
Nisan Stiennon, Long Ouyang, Jeffrey Wu, Daniel Ziegler, Ryan Lowe, Chelsea Voss, Alec Radford, Dario Amodei, and Paul~F Christiano.
\newblock Learning to summarize with human feedback.
\newblock {\em Advances in Neural Information Processing Systems}, 33:3008--3021, 2020.

\bibitem{christiano2017deeprlhf}
Paul~F Christiano, Jan Leike, Tom Brown, Miljan Martic, Shane Legg, and Dario Amodei.
\newblock Deep reinforcement learning from human preferences.
\newblock {\em Advances in neural information processing systems}, 30, 2017.

\bibitem{bai2022training}
Yuntao Bai, Andy Jones, Kamal Ndousse, Amanda Askell, Anna Chen, Nova DasSarma, Dawn Drain, Stanislav Fort, Deep Ganguli, Tom Henighan, et~al.
\newblock Training a helpful and harmless assistant with reinforcement learning from human feedback.
\newblock {\em arXiv preprint arXiv:2204.05862}, 2022.

\bibitem{rafailov2024DPO}
Rafael Rafailov, Archit Sharma, Eric Mitchell, Christopher~D Manning, Stefano Ermon, and Chelsea Finn.
\newblock Direct preference optimization: Your language model is secretly a reward model.
\newblock {\em Advances in Neural Information Processing Systems}, 36, 2024.

\bibitem{guo2024dap}
Shangmin Guo, Biao Zhang, Tianlin Liu, Tianqi Liu, Misha Khalman, Felipe Llinares, Alexandre Rame, Thomas Mesnard, Yao Zhao, Bilal Piot, Johan Ferret, and Mathieu Blondel.
\newblock Direct language model alignment from online ai feedback, 2024.

\bibitem{kirkpatrick2017overcoming}
James Kirkpatrick, Razvan Pascanu, Neil Rabinowitz, Joel Veness, Guillaume Desjardins, Andrei~A Rusu, Kieran Milan, John Quan, Tiago Ramalho, Agnieszka Grabska-Barwinska, et~al.
\newblock Overcoming catastrophic forgetting in neural networks.
\newblock {\em Proceedings of the national academy of sciences}, 114(13):3521--3526, 2017.

\bibitem{rolnick2019experience}
David Rolnick, Arun Ahuja, Jonathan Schwarz, Timothy Lillicrap, and Gregory Wayne.
\newblock Experience replay for continual learning.
\newblock {\em Advances in neural information processing systems}, 32, 2019.

\bibitem{zhang2024cppo}
Han Zhang, Yu~Lei, Lin Gui, Min Yang, Yulan He, Hui Wang, and Ruifeng Xu.
\newblock {CPPO}: Continual learning for reinforcement learning with human feedback.
\newblock In {\em The Twelfth International Conference on Learning Representations}, 2024.

\bibitem{lee2023rlaif}
Harrison Lee, Samrat Phatale, Hassan Mansoor, Kellie Lu, Thomas Mesnard, Colton Bishop, Victor Carbune, and Abhinav Rastogi.
\newblock Rlaif: Scaling reinforcement learning from human feedback with ai feedback.
\newblock {\em arXiv preprint arXiv:2309.00267}, 2023.

\bibitem{bolnick2001intraspecific}
Daniel~I Bolnick.
\newblock Intraspecific competition favours niche width expansion in drosophila melanogaster.
\newblock {\em Nature}, 410(6827):463--466, 2001.

\bibitem{thorne2003evolution}
Barbara~L Thorne, Nancy~L Breisch, and Mario~L Muscedere.
\newblock Evolution of eusociality and the soldier caste in termites: influence of intraspecific competition and accelerated inheritance.
\newblock {\em Proceedings of the National Academy of Sciences}, 100(22):12808--12813, 2003.

\bibitem{agarwal2019online}
Naman Agarwal, Brian Bullins, Elad Hazan, Sham Kakade, and Karan Singh.
\newblock Online control with adversarial disturbances.
\newblock In {\em International Conference on Machine Learning}, pages 111--119. PMLR, 2019.

\bibitem{hazan2020nonstochastic}
Elad Hazan, Sham Kakade, and Karan Singh.
\newblock The nonstochastic control problem.
\newblock In {\em Algorithmic Learning Theory}, pages 408--421. PMLR, 2020.

\bibitem{goodfellow2020generative}
Ian Goodfellow, Jean Pouget-Abadie, Mehdi Mirza, Bing Xu, David Warde-Farley, Sherjil Ozair, Aaron Courville, and Yoshua Bengio.
\newblock Generative adversarial networks.
\newblock {\em Communications of the ACM}, 63(11):139--144, 2020.

\bibitem{hu2021lora}
Edward~J Hu, Yelong Shen, Phillip Wallis, Zeyuan Allen-Zhu, Yuanzhi Li, Shean Wang, Lu~Wang, and Weizhu Chen.
\newblock Lora: Low-rank adaptation of large language models.
\newblock {\em arXiv preprint arXiv:2106.09685}, 2021.

\bibitem{lake2023human}
Brenden~M Lake and Marco Baroni.
\newblock Human-like systematic generalization through a meta-learning neural network.
\newblock {\em Nature}, 623(7985):115--121, 2023.

\bibitem{van2020brain}
Gido~M van~de Ven, Hava~T Siegelmann, and Andreas~S Tolias.
\newblock Brain-inspired replay for continual learning with artificial neural networks.
\newblock {\em Nature Communications}, 11(1), 2020.

\bibitem{abs-2403-02628}
Biqing Qi, Xingquan Chen, Junqi Gao, Dong Li, Jianxing Liu, Ligang Wu, and Bowen Zhou.
\newblock Interactive continual learning: Fast and slow thinking.
\newblock {\em CoRR}, abs/2403.02628, 2024.

\bibitem{jiang2024chasing}
Zhimeng~Stephen Jiang, Xiaotian Han, Hongye Jin, Guanchu Wang, Rui Chen, Na~Zou, and Xia Hu.
\newblock Chasing fairness under distribution shift: A model weight perturbation approach.
\newblock {\em Advances in Neural Information Processing Systems}, 36, 2024.

\bibitem{wallace2023diffusion}
Bram Wallace, Meihua Dang, Rafael Rafailov, Linqi Zhou, Aaron Lou, Senthil Purushwalkam, Stefano Ermon, Caiming Xiong, Shafiq Joty, and Nikhil Naik.
\newblock Diffusion model alignment using direct preference optimization, 2023.

\bibitem{pappa2024modipo}
Massimiliano Pappa, Luca Collorone, Giovanni Ficarra, Indro Spinelli, and Fabio Galasso.
\newblock Modipo: text-to-motion alignment via ai-feedback-driven direct preference optimization.
\newblock {\em arXiv preprint arXiv:2405.03803}, 2024.

\bibitem{majumder2024tango}
Navonil Majumder, Chia-Yu Hung, Deepanway Ghosal, Wei-Ning Hsu, Rada Mihalcea, and Soujanya Poria.
\newblock Tango 2: Aligning diffusion-based text-to-audio generations through direct preference optimization.
\newblock {\em arXiv preprint arXiv:2404.09956}, 2024.

\bibitem{zhang2024direct}
Ruohong Zhang, Liangke Gui, Zhiqing Sun, Yihao Feng, Keyang Xu, Yuanhan Zhang, Di~Fu, Chunyuan Li, Alexander Hauptmann, Yonatan Bisk, et~al.
\newblock Direct preference optimization of video large multimodal models from language model reward.
\newblock {\em arXiv preprint arXiv:2404.01258}, 2024.

\bibitem{yang2023direct}
Guangyu Yang, Jinghong Chen, Weizhe Lin, and Bill Byrne.
\newblock Direct preference optimization for neural machine translation with minimum bayes risk decoding.
\newblock {\em arXiv preprint arXiv:2311.08380}, 2023.

\bibitem{zhou2023onepreferencefitsall}
Zhanhui Zhou, Jie Liu, Chao Yang, Jing Shao, Yu~Liu, Xiangyu Yue, Wanli Ouyang, and Yu~Qiao.
\newblock Beyond one-preference-fits-all alignment: Multi-objective direct preference optimization, 2023.

\bibitem{pal2024smaug}
Arka Pal, Deep Karkhanis, Samuel Dooley, Manley Roberts, Siddartha Naidu, and Colin White.
\newblock Smaug: Fixing failure modes of preference optimisation with dpo-positive.
\newblock {\em arXiv preprint arXiv:2402.13228}, 2024.

\bibitem{xie2022general}
Jiangwei Xie, Shipeng Yan, and Xuming He.
\newblock General incremental learning with domain-aware categorical representations.
\newblock In {\em Proceedings of the IEEE/CVF Conference on Computer Vision and Pattern Recognition}, pages 14351--14360, 2022.

\bibitem{Simon_2022_CVPR}
Christian Simon, Masoud Faraki, Yi-Hsuan Tsai, Xiang Yu, Samuel Schulter, Yumin Suh, Mehrtash Harandi, and Manmohan Chandraker.
\newblock On generalizing beyond domains in cross-domain continual learning.
\newblock In {\em Proceedings of the IEEE/CVF Conference on Computer Vision and Pattern Recognition (CVPR)}, pages 9265--9274, June 2022.

\bibitem{wang2024comprehensive}
Liyuan Wang, Xingxing Zhang, Hang Su, and Jun Zhu.
\newblock A comprehensive survey of continual learning: Theory, method and application.
\newblock {\em IEEE Transactions on Pattern Analysis and Machine Intelligence}, 2024.

\bibitem{aljundi2018memory}
Rahaf Aljundi, Francesca Babiloni, Mohamed Elhoseiny, Marcus Rohrbach, and Tinne Tuytelaars.
\newblock Memory aware synapses: Learning what (not) to forget.
\newblock In {\em Proceedings of the European conference on computer vision (ECCV)}, pages 139--154, 2018.

\bibitem{chaudhry2018riemannian}
Arslan Chaudhry, Puneet~K Dokania, Thalaiyasingam Ajanthan, and Philip~HS Torr.
\newblock Riemannian walk for incremental learning: Understanding forgetting and intransigence.
\newblock In {\em Proceedings of the European conference on computer vision (ECCV)}, pages 532--547, 2018.

\bibitem{li2017learning}
Zhizhong Li and Derek Hoiem.
\newblock Learning without forgetting.
\newblock {\em IEEE transactions on pattern analysis and machine intelligence}, 40(12):2935--2947, 2017.

\bibitem{castro2018end}
Francisco~M Castro, Manuel~J Mar{\'\i}n-Jim{\'e}nez, Nicol{\'a}s Guil, Cordelia Schmid, and Karteek Alahari.
\newblock End-to-end incremental learning.
\newblock In {\em Proceedings of the European conference on computer vision (ECCV)}, pages 233--248, 2018.

\bibitem{sun2019lamol}
Fan-Keng Sun, Cheng-Hao Ho, and Hung-Yi Lee.
\newblock Lamol: Language modeling for lifelong language learning.
\newblock {\em arXiv preprint arXiv:1909.03329}, 2019.

\bibitem{liu2020generative}
Xialei Liu, Chenshen Wu, Mikel Menta, Luis Herranz, Bogdan Raducanu, Andrew~D Bagdanov, Shangling Jui, and Joost~van de~Weijer.
\newblock Generative feature replay for class-incremental learning.
\newblock In {\em Proceedings of the IEEE/CVF Conference on Computer Vision and Pattern Recognition Workshops}, pages 226--227, 2020.

\bibitem{abs-2403-04140}
Biqing Qi, Junqi Gao, Xingquan Chen, Dong Li, Jianxing Liu, Ligang Wu, and Bowen Zhou.
\newblock Contrastive augmented graph2graph memory interaction for few shot continual learning.
\newblock {\em CoRR}, abs/2403.04140, 2024.

\bibitem{pmlr-v151-rosenfeld22a}
Elan Rosenfeld, Pradeep Ravikumar, and Andrej Risteski.
\newblock An online learning approach to interpolation and extrapolation in domain generalization.
\newblock In Gustau Camps-Valls, Francisco J.~R. Ruiz, and Isabel Valera, editors, {\em Proceedings of The 25th International Conference on Artificial Intelligence and Statistics}, volume 151 of {\em Proceedings of Machine Learning Research}, pages 2641--2657. PMLR, 28--30 Mar 2022.

\bibitem{volpi2021continual}
Riccardo Volpi, Diane Larlus, and Gr{\'e}gory Rogez.
\newblock Continual adaptation of visual representations via domain randomization and meta-learning.
\newblock In {\em Proceedings of the IEEE/CVF Conference on Computer Vision and Pattern Recognition}, pages 4443--4453, 2021.

\bibitem{kundu2020class}
Jogendra~Nath Kundu, Rahul~Mysore Venkatesh, Naveen Venkat, Ambareesh Revanur, and R~Venkatesh Babu.
\newblock Class-incremental domain adaptation.
\newblock In {\em Computer Vision--ECCV 2020: 16th European Conference, Glasgow, UK, August 23--28, 2020, Proceedings, Part XIII 16}, pages 53--69. Springer, 2020.

\bibitem{zhao2021improved}
Peng Zhao and Lijun Zhang.
\newblock Improved analysis for dynamic regret of strongly convex and smooth functions.
\newblock In {\em Learning for Dynamics and Control}, pages 48--59. PMLR, 2021.

\bibitem{haghtalab2020smoothed}
Nika Haghtalab, Tim Roughgarden, and Abhishek Shetty.
\newblock Smoothed analysis of online and differentially private learning.
\newblock {\em Advances in Neural Information Processing Systems}, 33:9203--9215, 2020.

\bibitem{radford2019language}
Alec Radford, Jeffrey Wu, Rewon Child, David Luan, Dario Amodei, Ilya Sutskever, et~al.
\newblock Language models are unsupervised multitask learners.
\newblock {\em OpenAI blog}, 1(8):9, 2019.

\bibitem{ziegler2019fine}
Daniel~M Ziegler, Nisan Stiennon, Jeffrey Wu, Tom~B Brown, Alec Radford, Dario Amodei, Paul Christiano, and Geoffrey Irving.
\newblock Fine-tuning language models from human preferences.
\newblock {\em arXiv preprint arXiv:1909.08593}, 2019.

\bibitem{gpt-j}
Ben Wang and Aran Komatsuzaki.
\newblock {GPT-J-6B: A 6 Billion Parameter Autoregressive Language Model}.
\newblock \url{https://github.com/kingoflolz/mesh-transformer-jax}, May 2021.

\bibitem{chen2023exploring}
Yi~Chen, Rui Wang, Haiyun Jiang, Shuming Shi, and Ruifeng Xu.
\newblock Exploring the use of large language models for reference-free text quality evaluation: A preliminary empirical study.
\newblock {\em arXiv preprint arXiv:2304.00723}, 2023.

\bibitem{llama3modelcard}
AI@Meta.
\newblock Llama 3 model card.
\newblock 2024.

\bibitem{gao2023scaling}
Leo Gao, John Schulman, and Jacob Hilton.
\newblock Scaling laws for reward model overoptimization.
\newblock In {\em International Conference on Machine Learning}, pages 10835--10866. PMLR, 2023.

\bibitem{lin2004rouge}
Chin-Yew Lin.
\newblock Rouge: A package for automatic evaluation of summaries.
\newblock In {\em Text summarization branches out}, pages 74--81, 2004.

\bibitem{Aljundi_2019_CVPR}
Rahaf Aljundi, Klaas Kelchtermans, and Tinne Tuytelaars.
\newblock Task-free continual learning.
\newblock In {\em Proceedings of the IEEE/CVF Conference on Computer Vision and Pattern Recognition (CVPR)}, June 2019.

\bibitem{kaplanis2019policy}
Christos Kaplanis, Murray Shanahan, and Claudia Clopath.
\newblock Policy consolidation for continual reinforcement learning.
\newblock {\em arXiv preprint arXiv:1902.00255}, 2019.

\bibitem{schopf2022hypernetwork}
Philemon Sch{\"o}pf, Sayantan Auddy, Jakob Hollenstein, and Antonio Rodriguez-Sanchez.
\newblock Hypernetwork-ppo for continual reinforcement learning.
\newblock In {\em Deep Reinforcement Learning Workshop NeurIPS 2022}, 2022.

\bibitem{Ramamurthy2022IsRL}
Rajkumar Ramamurthy, Prithviraj Ammanabrolu, Kiant{\'e} Brantley, Jack Hessel, Rafet Sifa, Christian Bauckhage, Hannaneh Hajishirzi, and Yejin Choi.
\newblock Is reinforcement learning (not) for natural language processing?: Benchmarks, baselines, and building blocks for natural language policy optimization.
\newblock 2022.

\bibitem{freund1997decision}
Yoav Freund and Robert~E Schapire.
\newblock A decision-theoretic generalization of on-line learning and an application to boosting.
\newblock {\em Journal of computer and system sciences}, 55(1):119--139, 1997.

\end{thebibliography}
\clearpage
\pagenumbering{arabic}
\appendix
\section{Mathematical Derivations}
\subsection{The Proof of Lemma \ref{boundlem}}\label{proofboundlem}
\begin{proof}

To derive our theorem, we first present a necessary lemma on regret bounds.
\begin{lemma}\label{othersbound}\cite{haghtalab2020smoothed}
    Consider an algorithm $\mathcal{A}$ that uses Hedge \cite{freund1997decision} on a finite hypothesis class $\mathcal H'$ instead of $\mathcal H$, with the expected regret is defined as
    \begin{equation}
        \mathbb{E}[\operatorname{REGRET}(\mathcal{A}, \mathcal{D})]=\frac{1}{T}\mathbb{E}_{\mathbf{s} \sim \mathcal{D}}\left[\sum_{t=1}^T \operatorname{err}_{s_t}\left(h_t\right)-\min _{h \in H} \sum_{t=1}^T \operatorname{err}_{s_t}(h)\right],
    \end{equation}
    where $\operatorname{err}_s(h)=\mathbbm{1}_{\{h(s)\ne y\}}$. The expected regret has the upper bound below:
    \begin{equation}
        \mathbb{E}[\operatorname{REGRET}(\mathcal{A}, \mathcal{D})] \leq O\left(\sqrt{ \ln \left(\left|\mathcal{H}^{\prime}\right|\right)/T}\right)+\frac{1}{T}\mathbb{E}_{\mathbf{s}\sim\mathcal{D}}\left[\max _{h \in \mathcal{H}} \min _{h^{\prime} \in \mathcal{H}^{\prime}} \sum_{t=1}^T \mathbbm{1}_{\{h\left(s_t\right) \neq h^{\prime}\left(s_t\right)\}}\right].
    \end{equation} 
\end{lemma}

Note that our regret \( R(T) \) differs from \(\operatorname{REGRET}(\mathcal{A}, \mathcal{D})\) in the aforementioned theorem by at most a constant factor of \(\ln 2\), which represents the maximum value of \( l(\theta, x) \). Therefore, the term $O\left(\sqrt{\ln\left(\left|\mathcal{H}^{\prime}\right|\right)/T}\right)$ in eq. (15) still holds. Leveraging Lemma \ref{othersbound}, we can derive the following result:
\begin{equation}
    \begin{split}
        R(T)=&\mathbb{E} _{x\sim \mathcal D}{\frac{1}{T}\big[\sum_{t=1}^{T}l(h_t,x_t)- \text{min}_{h\in \mathcal H} \sum_{t=1}^{T} l(h,x_t)\big]}\\
        =&\mathbb{E} _{x\sim \mathcal D}{\frac{1}{T}\big[\sum_{t=1}^{T}l(h_t,x_t)- \text{min}_{h'\in \mathcal H'} \sum_{t=1}^{T} l(h',x_t)\big]}\\
        &+\frac{1}{T}\big[\text{min}_{h'\in \mathcal H'}\sum_{t=1}^{T}l(h'_t,x_t)- \text{min}_{h\in \mathcal H} \sum_{t=1}^{T} l(h,x_t)\big]\\
        \le& O(\sqrt{\text{ln}(\mathcal H')/T}) + \mathbb{E} _{x\sim \mathcal D}\frac{1}{T}\big[\text{min}_{h'\in \mathcal H'}\text{max}_{h\in \mathcal H}(\sum_{t=1}^{T}l(h'_t,x_t)- \sum_{t=1}^{T}l(h,x_t))\big].
    \end{split}
\end{equation}
\end{proof}

\subsection{The Proof of Theorem \ref{singletaskbound}}\label{proofofolfs}

\begin{proof}

    \textbf{Vanilla DPO:}
    \begin{equation}\label{eq17}
		\begin{split}
            l(\theta_{t-k},x_{t-k})-l(\theta_t,x_t) &=	l(\theta_{t-k},x_{t-k})-l(\theta_{t},x_{t-k}) 
            + l(\theta_{t},x_{t-k}) -	l(\theta_t,x_t) \\
            &\le g_{t-k}(\theta_{t-k}-\theta_t)  +  l(\theta_{t},x_{t-k}) -	l(\theta_t,x_t),
        \end{split}
    \end{equation}
    where $k$ is a positive integer. According to assumptions \ref{gb} and \ref{pb}, we have $|g_{t-k}(\theta_{t-k}-\theta_t)| \le Gd;\ l(\theta_T,x_i)-l(\theta_i,x_i) \ge - Gd$, on the other hand,
    \begin{equation}
        l(\theta_{t},x_{t-k}) - l(\theta_t,x_t) = l(\theta_{t},x_{t-k})-\mathbb E[l(\theta_t)] + \mathbb E[l(\theta_t)] - l(\theta_t,x_t),
    \end{equation}
    where $l(\theta_t,x_1),l(\theta_t,x_2),\dots,l(\theta_t,x_T)  \underset{iid.}{\sim} l(\theta_t)$, 
    $l(\theta_t)$ is a distribution of objective function conditioned on the parameters $\theta_t$. 
    According to the Hoeffding's inequality, $\exists\ \delta_0\in(0,1)$ s.t.
    \begin{equation}
        P\left(l(\theta_{t},x_{t-k})-\mathbb E[l(\theta_t)] \le \text{ln}2\sqrt{-\frac{\text{ln}\delta_0}{2}}\right) \ge 1- \delta_0,
    \end{equation}
    likewise, we have
    \begin{equation}
        P\left(\mathbb E[l(\theta_t)]-l(\theta_{t},x_{t}) \ge \text{ln}2\sqrt{-\frac{\text{ln}\delta_0}{2}} \right) \le \delta_0,         
    \end{equation}
    which equals to
    \begin{equation}
        \ P\left(l(\theta_{t},x_{t})-\mathbb E[l(\theta_t)] \le \text{ln}2\sqrt{-\frac{\text{ln}\delta_0}{2}} \right) \ge 1-\delta_0.
    \end{equation}
    substituting into the right side of eq. (17), we have
    \begin{equation}
    l(\theta_{t-k},x_{t-k}) - l(\theta_t,x_t) \le Gd +2\text{ln}2\sqrt{-\frac{\text{ln}\delta_0}{2}}
    \end{equation}
    holds with probability $ (1- \delta_0)^2$. Furthermore, we can obtain:
    \begin{equation}
        l(\theta_i,x_i)  \ge  l(\theta_1,x_1)-(Gd +2\text{ln}2\sqrt{-\frac{\text{ln}\delta}{2}}).
    \end{equation}
    
    Denote $\delta=2(T-1)\delta_0- (T-1)(2T-3)(\delta_0)^2[1-\delta_0]^{2T-4}$, we have the following inequality holds with probability $1-\delta$: 
    \begin{equation}
        \begin{split}
            R(T) &= \frac{1}{T}\big[\sum_{i=1}^{T} [l(\theta_T,x_i)- l(\theta_i,x_i)+l(\theta_i,x_i)- l(\theta^*,x_i)]\big]\\
            &\ge  -Gd + l(\theta_1,x_1)-\frac{1}{T}\sum_{i=1}^{T}l(\theta^*,x_i)-(1-\frac{1}{T})(Gd+2\text{ln}2\sqrt{-\frac{\text{ln}\delta_0}{2}})\\
            &= l(\theta_1,x_1)-\frac{1}{T}\sum_{i=1}^{T}l(\theta^*,x_i)-(2-\frac{1}{T})Gd-2(1-\frac{1}{T})\text{ln}2\sqrt{-\frac{\text{ln}\delta_0}{2}}.
        \end{split}       
    \end{equation}
    After introducing \textbf{fast and slow modules}:
    \begin{equation}\label{eq24}
        \begin{split}
            &l(\theta_{s-k},x_{s-k})- 	l(\theta_s,x_s)\\
            \le &	l(\theta_{s-k},x_{s-k})-	l(\theta_s,x_{s-k}) + 	l(\theta_s,x_{s-k}) -	\text{min} (l(\theta_s,x_s),l(w_s,x_s)) \\
            \le & g_{s-k}(\theta_{s-k}-\theta_s)  +  
            l(\theta_s,x_{s-k})
            -\frac{l(\theta_s,x_s)+l(w_s,x_s)-|l(\theta_s,x_s)-l(w_s,x_s)|}{2}\\
            \le & Gd 
            +l(\theta_s,x_{s-k})-l(\theta_s,x_s)
            +\frac{l(\theta_s,x_s)-l(w_s,x_s)}{2}
            +\frac{|l(\theta_s,x_s)-l(w_s,x_s)|}{2}\\
            \le & l(\theta_s,x_{s-k})-l(\theta_s,x_s)+ 2 G d .
        \end{split}
    \end{equation}
    Similar to the process in eq. (18) - eq. (23), we have the derivations from eq. (26) - eq. (30):
    \begin{equation}
        l(\theta_{s},x_{s-k}) - l(\theta_s,x_s) = l(\theta_{s},x_{s-k})-\mathbb E[l(\theta_s)] + \mathbb E[l(\theta_s)] - l(\theta_s,x_s),
    \end{equation}
    where $l(\theta_s,x_1),l(\theta_s,x_2),\dots,l(\theta_s,x_T)\  \underset{iid.}{\sim} l(\theta_s)$,
    $l(\theta_s)$ is a distribution of objective function conditioned on the parameters $\theta_s$. 
    According to Hoeffding's inequality,
    \begin{equation}
        P\left(l(\theta_{s},x_{s-k})-\mathbb E[l(\theta_s)] \le \text{ln}2\sqrt{-\frac{\text{ln}\delta_0}{2}}\right) \ge 1- \delta_0,
    \end{equation}
    \begin{equation}
        P\left(\mathbb E[l(\theta_s)]-l(\theta_{s},x_{s}) \ge \text{ln}2\sqrt{-\frac{\text{ln}\delta_0}{2}} \right) \le \delta_0,     
    \end{equation}
    \begin{equation}
        \text{thus\ }\ P\left(l(\theta_{s},x_{s})-\mathbb E[l(\theta_s)] \le \text{ln}2\sqrt{-\frac{\text{ln}\delta_0}{2}} \right) \ge 1-\delta_0.
    \end{equation}

    Hence, we can draw a similar conclusion.
    \begin{equation}
	    P\left[l(\theta_s,x_{s-k})-l(\theta_s,x_s)\le 2\text{ln}2\sqrt{-\frac{\text{ln}\delta_0}{2}}\right]  \ge (1- \delta_0)^2.
    \end{equation}
    Therefore, in the setting of OFS-DPO, the eq. (25) gives $l(\theta_{s-k},x_{s-k})-l(\theta_s,x_s) \le 2Gd+2c$ with probability $ (1- \delta_0)^2$. Then the follwing equation establishes with probability $1-\delta$
    \begin{equation}
        \hat R(T)  \ge  l(\theta_1,x_1)-\frac{1}{T}\sum_{i=1}^{T}l(\theta^*,x_i)-(3-\frac{1}{T})Gd-2(1-\frac{1}{T})\text{ln}2\sqrt{-\frac{\text{ln}\delta_0}{2}}.
    \end{equation}
    Hence the OFS-DPO algorithm has a smaller lower bound of regret than vanilla DPO.
\end{proof}

\subsection{The Proof of Proposition \ref{stablegradients}}
\begin{proof}
    Consider the objective function of conventional DPO method:
    \begin{equation}
        \mathcal L_{DPO}(\theta)=-\mathbb E_{(x,y_w,y_l) \sim \mathcal D}[\text{log} \sigma(\beta \text{log} \frac{\pi_{\theta}(y_w|x)}{\pi_{ref}(y_w|x)}-\beta \text{log} \frac{\pi_{\theta}(y_l|x)}{\pi_{ref}(y_l|x)})],
    \end{equation}
    where its gradient is 
    \begin{equation}
        \begin{split}
    &\nabla_\theta \mathcal L_{DPO}(\theta)=\\
    &-\beta \mathbb E_{(x,y_w,y_l) \sim \mathcal D}
    [\sigma(\hat{r}_\theta(x,y_l)-\hat{r}_\theta(x,y_w))][\nabla_\theta \text{log}\pi_{\theta}(y_w|x)  -\nabla_\theta \text{log}\pi_{\theta}(y_l|x)],
        \end{split}
    \end{equation}
    where $\hat{r}_\theta(x,y)\triangleq\beta \frac{\pi_{\theta}(y|x)}{\pi_{ref}(y|x)}$.

    \textbf{(\romannumeral1)In DPO method}, $\pi_{\theta}(y_l|x)\rightarrow 0$,  $\pi_{\theta}(y_w|x)\rightarrow 1$. Therefore, $\forall \epsilon >0, \exists N>0, C>0$, and $N+C>\frac{\text{ln}(1-\epsilon/\epsilon)}{\beta}$, s.t.,
    \begin{equation}
        \begin{split}
    &\text{log}\frac{\pi_{\theta}(y_l|x)}{\pi_{ref}(y_l|x)} <-N,\\
    &\text{log}\frac{\pi_{\theta}(y_w|x)}{\pi_{ref}(y_w|x)} >C.
        \end{split}
    \end{equation}
    Note that the coefficient of DPO gradient tends towards an exceedingly small value, i.e.,
    \begin{equation}
        \begin{split}
            \sigma(\hat{r}_\theta(x,y_l)-\hat{r}_\theta(x,y_w))\le\sigma(\beta(-N-C))=\frac{1}{1+\text{e}^{\beta(N+C)}}
            \le\epsilon.
        \end{split}
    \end{equation}

    \textbf{(\romannumeral2)In OFS-DPO method}, $\mathcal L_{DPO-FS}$ is incorporated as a regularization term, influencing the acquired gradients during model iteration.
    \begin{equation}
        \mathcal L_{DPO-new}(\theta^F)=\mathcal L_{DPO}(\theta^F)+\alpha \mathcal L_{DPO-FS}.
    \end{equation}
    Its gradient can be calculated as follows: 
    \begin{equation} \label{grapt}
        \begin{split}
            &\nabla_{\theta^F}	\mathcal L_{DPO-new}(\theta^F)\\
            =&\underbrace{-\beta \mathbb E_{(x,y_w,y_l) \sim \mathcal D}
    ( \sigma(\hat{r}_{\theta^F}(x,y_l)-\hat{r}_{\theta^F}(x,y_w))[\nabla_{\theta^F} \text{log}\pi_{\theta^F}(y_w|x)  -\nabla_{\theta^F} \text{log}\pi_{\theta^F}(y_l|x)]}_{first\ term}\\
            & \underbrace{+2\alpha\sigma(\hat{r}_{FS}(x,y_l)-\hat{r}_{FS}(x,y_w))[\nabla_{\theta^F} \text{log}\pi_{\theta^F}(y_w|x) -\nabla_{\theta^F} \text{log}\pi_{\theta^F}(y_l|x)]}_{second\ term},
        \end{split}
    \end{equation}
    where $\hat{r}_{\theta^F}(x,y)\triangleq\frac{\pi_{\theta^F}(y|x)}{\pi_{ref}(y|x)}$,  $\hat{r}_{FS}(x,y)\triangleq\frac{\pi_{\theta^F}(y|x)}{\pi_{\theta^S}(y|x)}$.
    
    The first term of the eq. (37) is equivalent to the form of the original DPO objective function. As elucidated in the analysis of \textbf{DPO method}, this term gradually diminishes towards zero with the proceed of training, i.e.,$\sigma(\hat{r}_\theta(x,y_l)-\hat{r}_\theta(x,y_w) )
    \le \epsilon$.

    In the training stage, F-module and S-module converge to the same objective: $\pi_{\theta^F}(y_l|x)\rightarrow 0$,  $\pi_{\theta^F}(y_w|x)\rightarrow 1$, $\pi_{\theta^S}(y_l|x)\rightarrow 0$,  $\pi_{\theta^S}(y_w|x)\rightarrow 1$.  For $\epsilon>0$, $\exists\ \delta_1>0$, s.t.,
    $\pi_{\theta^F}(y_w|x)-\pi_{\theta^S}(y_w|x)=\pi_{\theta^S}(y_l|x)-\pi_{\theta^F}(y_l|x) < \delta_1$. 

    Choose $\delta_1=\pi_{\theta^S}(y_l|x)(1- \text{e}^{\text{ln}2+\frac{\text{ln}(\epsilon/1-\epsilon)}{\beta}}) $, and there are the following results:
    \begin{equation}
        \begin{split}
        &\frac{\pi_{\theta^F}(y_w|x)}{\pi_{\theta^S}(y_w|x)} \in (1,2)\\
        \Rightarrow&\frac{\pi_{\theta^F}(y_l|x)}{\pi_{\theta^S}(y_l|x)} +\frac{\delta_1}{\pi_{\theta^S}(y_l|x)} >1  \\
        \Rightarrow&\frac{\pi_{\theta^F}(y_l|x)}{\pi_{\theta^S}(y_l|x)} > \text{e}^{\text{ln}2+\frac{\text{ln}(\epsilon/(1-\epsilon))}{\beta}},
        \end{split}
    \end{equation}
    substituting into $\sigma(\hat{r}_{FS}(x,y_l)-\hat{r}_{FS}(x,y_w) )$, we have
    \begin{equation}
        \begin{split}
        &	\sigma(\hat{r}_{FS}(x,y_l)-\hat{r}_{FS}(x,y_w) )\\
        = &\sigma(\beta\text{log}\frac{\pi_{\theta^F}(y_l|x)}{\pi_{\theta^S}(y_l|x)}-\beta\text{log}\frac{\pi_{\theta^F}(y_w|x)}{\pi_{\theta^S}(y_w|x)})\\
    >&\sigma	(\text{ln}(\frac{\epsilon}{1-\epsilon}))\\
    =&\epsilon.
        \end{split}
    \end{equation}
\end{proof}

\subsection{The Proof of Theorem \ref{multitaskbound}}
\begin{proof}
    
    Before proceeding with our formal proof, we first present some necessary theorems \cite{jiang2024chasing}:
    \begin{theorem}\label{datadisshift}
        Given source and target datasets with probability distribution $\mathcal P_{\mathcal S}$ and $\mathcal P_{\mathcal T}$ , there exists data perturbation so that the training loss of any neural network $l(\theta,\cdot)$ for target distribution equals that for source distribution with data perturbation, i.e.,
        \begin{equation}
            \mathbb{E}_{x \sim \mathcal{P}_{\mathcal T}} \left[l\left(\theta,x\right)\right]=\mathbb{E}_{\delta_x(x)} \mathbb{E}_{x \sim \mathcal{P}_{\mathcal S}} \left[l\left(\theta,x+\delta_x(x)\right)\right].
        \end{equation}
    \end{theorem}
    \begin{theorem}\label{dataparashift}
       
        Considering the source dataset with distribution $\mathcal{P}_{\mathcal S}$, suppose the source dataset is perturbed with data perturbation $\delta$, and the loss of the neural network is given by $l(\theta,\cdot)$. In the general case, there exists a model weight perturbation $\Delta \theta$ such that the training loss on the perturbed source dataset is the same as the training loss with the model weight perturbation $\Delta \theta$ on the source distribution:
        \begin{equation}        \mathbb{E}_{\delta_x(x)}\mathbb{E}_{x\sim\mathcal{P}_{\mathcal S}}[l(\theta,x+\delta_x(x))]=\mathbb{E}_{x\sim\mathcal{P}_{\mathcal S}}[l(\theta+\Delta\theta,x)].
        \end{equation}
    \end{theorem}

    By the Definition \ref{dualtaskregret}, the empirical regret in cross-domain scenarios can be expressed as follows:
    \begin{equation}
    \begin{split}
        &R(T_1,T_2)\\
        =&[\frac{1}{T_1}\sum_{i=1}^{T_1}l(\theta_{T_1},s_i^{(1)})+ \frac{1}{T_2}\sum_{j=1}^{T_2}l(\theta_{T_2},s_j^{(2)})]-[\frac{1}{T_1}\sum_{i=1}^{T_1}l(\theta^*,s_i^{(1)})+ \frac{1}{T_2}\sum_{j=1}^{T_2}l(\theta^*,s_j^{(2)})]\\
        =&[\frac{1}{T_1}\sum_{i=1}^{T_1}l(\theta_{T_1},s_i^{(1)})+ \frac{1}{T_2}\sum_{j=1}^{T_2}l(\theta_{T_2},s_j^{(2)})]-[\frac{1}{T_1}\sum_{i=1}^{T_1}l(\theta(\beta),s_i^{(1)})+ \frac{1}{T_2}\sum_{j=1}^{T_2}l(\theta(\beta)),s_j^{(2)})]\\
        &+[\frac{1}{T_1}\sum_{i=1}^{T_1}l(\theta(\beta),s_i^{(1)})+ \frac{1}{T_2}\sum_{j=1}^{T_2}l(\theta(\beta),s_j^{(2)})]-[\frac{1}{T_1}\sum_{i=1}^{T_1}l(\theta^*,s_i^{(1)})+\frac{1}{T_2}\sum_{j=1}^{T_2}l(\theta^*,s_j^{(2)})]\\
        =&\underbrace{L(\theta_{T_1},\theta_{T_2})-L(\theta(\beta),\theta(\beta))}_{first\ term}+\underbrace{L(\theta(\beta),\theta(\beta))-L(\theta^*,\theta^*)}_{second\ term},
    \end{split}
    \end{equation}
    where $\theta(\beta) =\theta_0+\beta\Delta\theta_{(1)}+(1-\beta)\Delta\theta^{(2)}$. For the first term of $R(T_1,T_2)$, $L(\theta_{T_1},\theta_{T_2})-L(\theta(\beta),\theta(\beta))$, we have:
    \begin{equation}\label{eq36}
        \underbrace{L(\theta_{T_1},\theta_{T_2})-L(\theta^{(1)},\theta^{(2)})}_{first\ term}+\underbrace{L(\theta^{(1)},\theta^{(2)})-L(\theta(\beta),\theta(\beta))}_{second\ term}.
    \end{equation}
    The first term of eq. (43) can be written as :
    \begin{equation}
        \begin{split}
            A_1&\triangleq L(\theta_{T_1},\theta_{T_2})-L(\theta^{(1)},\theta^{(2)})\\
            &=\frac{1}{T_1}\sum_{i=1}^{T_1}(l(\theta_{T_1},s_i^{(1)})-l(\theta^{(1)},s_i^{(1)}))+ \frac{1}{T_2}\sum_{j=1}^{T_2}(l(\theta_{T_2},s_j^{(1)})-l(\theta^{(2)},s_j^{(2)}))].
        \end{split}
    \end{equation}
    Using the result of theorem \ref{singletaskbound}, there exist $\delta_1,\delta_2\in (0,1)$, s.t.
    \begin{equation}
    \begin{split}
        A_1 \ge& l(\theta_1,s_1^{(1)})-\frac{1}{T_1}\sum_{i=1}^{T_1}l(\theta^{(1)},s_i^{(1)})-\big[2-\frac{1}{T_1}+(1-\frac{1}{T_1})\mathbbm{1}_{\{mode=FS\}}\big]Gd\\
        &-2(1-\frac{1}{T_1})\ln2\sqrt{-\frac{\ln\delta_1}{2}}+l(\theta_1,s_1^{(2)})-\frac{1}{T_2}\sum_{j=1}^{T_2}l(\theta^{(2)},s_j^{(2)})\\
        &-\big[2-\frac{1}{T_2}+(1-\frac{1}{T_2})\mathbbm{1}_{\{mode=FS\}}\big]Gd-2(1-\frac{1}{T_2})\ln2\sqrt{-\frac{\ln\delta_2}{2}}\\
        \ge& l_1(s_1^{(1)},s_1^{(2)})-B(T_1,T_2)-[4-\frac{T_1+T_2}{T_1T_2}+(2-\frac{T_1+T_2}{T_1T_2})\mathbbm{1}_{\{mode=FS\}}]Gd\\
        &-2(1-\frac{T_1+T_2}{T_1T_2})c,
    \end{split}   
    \end{equation}
    here $B(T_1,T_2)=\frac{1}{T_1}\sum_{i=1}^{T_1}l(\theta^{(1)},s_{i}^{(1)})+\frac{1}{T_2}\sum_{j=1}^{T_2}l(\theta^{(2)},s_{j}^{(2)})$, $l_1(s_1^{(1)},s_1^{(2)})=l(\theta_1,s_1^{(1)})+l(\theta_1,s_1^{(2)})$, $c=\max\{\text{ln}2\sqrt{-\frac{\text{ln}\delta_1}{2}}$, $\text{ln}2\sqrt{-\frac{\text{ln}\delta_2}{2}}\}$,  $\delta_1,\delta_2\in (0,1)$ are constants, and $\mathbbm{1}_{\{mode=FS\}}$ represents whether to introduce fast and slow models.
    Meanwhile, the second term of eq. (43) can be denoted as $A_2$: 
    \begin{equation}
    \begin{split}
        A_2&\triangleq L(\theta^{(1)},\theta^{(2)})-L(\theta{(\beta)},\theta{(\beta)})\\
        &=\frac{1}{T_1}\sum_{i=1}^{T_1}(l(\theta^{(1)},s_i^{(1)})-l(\theta{(\beta)},s_i^{(1)}))+ \frac{1}{T_2}\sum_{j=1}^{T_2}(l(\theta{(\beta)},s_j^{(2)})-l(\theta^{(2)},s_j^{(2)})).
    \end{split}   
    \end{equation}
    
    For convenience of derivation, we make the following symbol conventions:
    \begin{equation}
        \delta\theta^{(1)}(\beta) \triangleq \Delta\theta^{(1)}-\Delta\theta(\beta)=(1-\beta)(\Delta\theta^{(1)}-\Delta\theta^{(2)}),
    \end{equation}
    \begin{equation}
        \delta\theta^{(2)}(\beta) \triangleq \Delta\theta^{(2)}-\Delta\theta(\beta)=\beta(\Delta\theta^{(2)}-\Delta\theta^{(1)}).
    \end{equation}
        
    According to theorem \ref{dataparashift} and theorem \ref{datadisshift},
    \begin{equation}\label{eq41}
        \mathbb{E}_{\delta_{\mathcal D_1}(x)}\mathbb{E}_{x\sim \mathcal D}l(\theta(\beta),x+\delta_{\mathcal D_1}(x))=\mathbb{E}_{x\sim \mathcal D}l(\theta(\beta)+\delta\theta^{(1)}(\beta),x).
    \end{equation}
    Perform the first-order Taylor expansion on both sides of the eq. (49), we have
    \begin{equation}
        \mathbb{E}_{\delta_{\mathcal D_1}(x)}\mathbb{E}_{x\sim \mathcal D}[l(\theta(\beta),x)+\nabla_x l(\theta(\beta),\hat{x}_1)\delta_{\mathcal D_1}(x)]=\mathbb{E}_{x\sim \mathcal D}[l(\theta(\beta),x)+\nabla_\theta l(\hat{\theta}^{(1)}(\beta),x)\delta\theta^{(1)}(\beta)].
    \end{equation}
    Hence, we can obtain
    \[
    \delta\theta^{(1)}(\beta)=(\mathbb{E}_{x\sim \mathcal D}\nabla_\theta l(\hat{\theta}^{(1)}(\beta),x))^{-1}\mathbb{E}_{\delta_{D_1}(x)}\mathbb{E}_{x\sim \mathcal D}\nabla_x l(\theta(\beta),\hat{x}_1)\delta_{D_1}(x),
    \]
    \[
    \delta\theta^{(2)}(\beta)=(\mathbb{E}_{x\sim \mathcal D}\nabla_\theta l(\hat{\theta}^{(2)}(\beta),x))^{-1}\mathbb{E}_{\delta_{D_2}(x)}\mathbb{E}_{x\sim \mathcal D}\nabla_x l(\theta(\beta),\hat{x}_2)\delta_{D_2}(x).
    \]
    And $A_2$ can be represented as:
    \begin{equation}
        A_2 = \frac{1}{T_1}\sum_{i=1}^{T_1}\nabla_\theta l(\hat{\theta}^{(1)}(\beta),s_i^{(1)})\delta\theta^{(1)}(\beta)+\frac{1}{T_2}\sum_{j=1}^{T_2}\nabla_\theta l(\hat{\theta}^{(2)}(\beta),s_j^{(2)})\delta\theta^{(2)}(\beta).
    \end{equation}
    Combine eq. (45) and eq. (51) and substitute into eq. (43), with the probility $(1-\delta_1)(1-\delta_2)$ holds:
    \begin{equation}
    \begin{split}
        A_1+A_2 \ge\  &l_1(s_1^{(1)},s_1^{(2)})-B(T_1,T_2)\\
        &-[4-\frac{T_1+T_2}{T_1T_2}+(2-\frac{T_1+T_2}{T_1T_2})\mathbbm{1}_{\{mode=FS\}}]Gd-2(1-\frac{T_1+T_2}{T_1T_2})c\\
        &+ \frac{1}{T_1}\sum_{i=1}^{T_1}\nabla_\theta l(\hat{\theta}^{(1)}(\beta),s_i^{(1)})\delta\theta^{(1)}(\beta)+\frac{1}{T_2}\sum_{j=1}^{T_2}\nabla_\theta l(\hat{\theta}^{(2)}(\beta),s_j^{(2)})\delta\theta^{(2)}(\beta).
    \end{split}
    \end{equation}
    For the second term of $R(T_1,T_2)\ i.e.\ L(\theta(\beta),\theta(\beta))-L(\theta^*,\theta^*)$, denote $\theta^*+\delta^{(1)}\triangleq\theta^{(1)},\ \theta^*+\delta^{(2)}\triangleq\theta^{(2)}\ (\theta_0+\Delta\theta^*+\delta^{(1)}=\theta_0+\Delta\theta^{(1)},\ \theta_0+\Delta\theta^*+\delta^{(2)}=\theta_0+\Delta\theta^{(2)})$, we make the derivations below:
    \begin{equation}
    \begin{split}
        L(\theta(\beta),\theta(\beta))-L(\theta^*,\theta^*)=&\frac{1}{T_1}\sum_{i=1}^{T_1}(l(\theta(\beta),s_i^{(1)})-l(\theta^*,s_i^{(1)}))\\
        &+ \frac{1}{T_2}\sum_{j=1}^{T_2}(l(\theta(\beta),s_j^{(2)})-l(\theta^*,s_j^{(2)}))\\
        =&[\frac{1}{T_1}\sum_{i=1}^{T_1}\nabla_\theta l(\hat{\theta}^{(1)}(\beta),s_i^{(1)})+\frac{1}{T_2}\sum_{j=1}^{T_2}\nabla_\theta l(\hat{\theta}^{(2)}(\beta),s_j^{(2)})](\theta(\beta)-\theta^*),
    \end{split}   
    \end{equation}
    where $\theta(\beta)-\theta^*=\beta\delta\theta^{(1)}+(1-\beta)\delta\theta^{(2)}$. Then combined with eq. (42), we obtain
    \begin{equation}
    \begin{split}
        R(T_1,T_2) \ge\  &l_1(s_1^{(1)},s_1^{(2)})-B(T_1,T_2)\\
        &-[4-\frac{T_1+T_2}{T_1T_2}+(2-\frac{T_1+T_2}{T_1T_2})\mathbbm{1}_{\{mode=FS\}}]Gd-2(1-\frac{T_1+T_2}{T_1T_2})c\\
        &+ \frac{1}{T_1}\sum_{i=1}^{T_1}\nabla_\theta l(\hat{\theta}^{(1)}(\beta),s_i^{(1)})\delta\theta^{(1)}(\beta)+\frac{1}{T_2}\sum_{j=1}^{T_2}\nabla_\theta l(\hat{\theta}^{(2)}(\beta),s_j^{(2)})\delta\theta^{(2)}(\beta)\\
        &+[\frac{1}{T_1}\sum_{i=1}^{T_1}\nabla_\theta l(\hat{\theta}^{(1)}(\beta),s_i^{(1)})+\frac{1}{T_2}\sum_{j=1}^{T_2}\nabla_\theta l(\hat{\theta}^{(2)}(\beta),s_j^{(2)})](\theta(\beta)-\theta^*).
    \end{split}
    \end{equation}
    For all summation terms over distribution $\mathcal D_1$, we can derive the lower bound below:
    \begin{equation}
        \begin{aligned} & 
        \sum_{i=1}^{T_1} \nabla_\theta l\left(\hat{\theta}^{(1)}(\beta), s_i^{(1)}\right) \delta\theta^{(1)}(\beta)+\sum_{i=1}^{T_1} \nabla_\theta l\left(\hat{\theta}^{(1)}, s_i^{(1)}\right)\left(\beta \delta \theta^{(1)}+(1-\beta) \delta^{(2)}\right) \\  =&\sum_{i=1}^{T_1} \nabla_\theta l\left(\hat{\theta}^{(1)}(\beta), s_i^{(1)}\right) \delta \theta^{(1 )}(\beta)-\sum_{i=1}^{T_1} \nabla_\theta l\left(\hat{\theta}^{(1)}, s_i^{(1)}\right) \delta \theta^{(1)}(\beta) \\ & +\sum_{i=1}^{T_1} \nabla_\theta l\left(\hat{\theta}^{(1)}, s_i^{(1)}\right) \delta \theta^{(1)}(\beta)+\sum_{i=1}^{T_1} \nabla_\theta l\left(\hat{\theta}^{(1)}, s_i^{(1)}\right)\left(\beta \delta \theta^{(1)}+(1-\beta) \delta \theta^{(2)}\right) \\  =&\sum_{i=1}^{T_1} \nabla_\theta l\left(\hat{\theta}^{(1)}(\beta), s_i^{(1)}\right) \delta \theta^{(1)}(\beta)-\sum_{i=1}^{T_1} \nabla_\theta l\left(\hat{\theta}^{(1)}, s_i^{(1)}\right) \delta \theta^{(l)}(\beta) \\ & +\sum_{i=1}^{T_1} \nabla_\theta l\left(\hat{\theta}^{(1)}, s_i^{(1)}\right)\left(\Delta \theta^{(1)}-\Delta \theta^*\right) \\  
        \ge&-T_1 G d+\sum_{i=1}^{T_1} \nabla_\theta l\left(\hat{\theta}^{(1)}, s_i^{(1)}\right)\left(\Delta \theta^{(1)}-\Delta \theta^*\right) \\ 
        \ge& -T_1 Gd.
        \end{aligned}
    \end{equation}
    The last inequality holds because $\nabla_\theta l(\hat{\theta}^{(1)}, s_i^{(1)}) (\Delta \theta^{(1)} - \Delta \theta^*)$ can be viewed as the inner product of two vectors. Specifically, $\nabla_\theta l(\hat{\theta}^{(1)}, s_i^{(1)}) (\Delta \theta^{(1)} - \Delta \theta^*) = m_1 m_2 \gamma$, where $m_1 = \|\nabla_\theta l(\hat{\theta}^{(1)}, s_i^{(1)})\|$, $m_2 = \|\Delta \theta^{(1)} - \Delta \theta^*\|$, and $\gamma$ represents the cosine of the angle between these two vectors. We note that $\nabla_\theta l(\hat{\theta}^{(1)}, s_i^{(1)})$ points to the optimal parameter on distribution $\mathcal D_1$, while $(\Delta \theta^{(1)} - \Delta \theta^*)$ represents the direction from the optimal parameter on the dual-task distribution to the optimal parameter on distribution $\mathcal D_1$. In theory, the angle between these two vectors is less than 90 degrees, i.e., $\nabla_\theta l(\hat{\theta}^{(1)}, s_i^{(1)}) (\Delta \theta^{(1)} - \Delta \theta^*) > 0$. Similarly, there are analogous properties for all summation terms over distribution $D_2$ that ensure $\sum_{j=1}^{T_2} \nabla_\theta l\left(\hat{\theta}^{(2)}(\beta), s_j^{(2)}\right) \delta \theta^{(2)}(\beta) + \sum_{j=1}^{T_2} \nabla_\theta l\left(\hat{\theta}^{(2)}, s_j^{(2)}\right)\left(\beta \delta \theta^{(2)} + (1-\beta) \delta \theta^{(2)}\right) \ge -T_2 G d$ holds true. Hence,
    \begin{equation}
        \begin{split}
        R(T_1,T_2) \ge\  &l_1(s_1^{(1)},s_1^{(2)})-B(T_1,T_2)\\
        &-[4-\frac{T_1+T_2}{T_1T_2}+(2-\frac{T_1+T_2}{T_1T_2})\mathbbm{1}_{\{mode=FS\}}]Gd-2(1-\frac{T_1+T_2}{T_1T_2})c\\
        &- 2Gd\\
        =& l_1(s_1^{(1)},s_1^{(2)})-B(T_1,T_2)\\
        &-[6-\frac{T_1+T_2}{T_1T_2}+(2-\frac{T_1+T_2}{T_1T_2})\mathbbm{1}_{\{mode=FS\}}]Gd-2(1-\frac{T_1+T_2}{T_1T_2})c.
    \end{split}
    \end{equation}
\end{proof}
\clearpage
\section{Proposed Algorithms}
\label{proposed algorithms}

The OFS-DPO algorithm is presented in Algorithm \ref{algo1}. Unlike DPO, we incorporate fast and slow modules to simulate intraspecific competition, thereby accelerating the evolutionary process. The COFS-DPO algorithm is depicted in Algorithm \ref{algo2}. For task-1 in the cross-domain setting, the parameters of the fast module, \(\theta_{1}^{F}\), are trained and subsequently used to initialize both the fast and slow modules for task-2. Training then proceeds on task-2. Ultimately, the parameters of the fast modules from both tasks, \(\theta_{1}^{F}\) and \(\theta_{2}^{F}\), are combined with appropriate weights to form the final model.
\begin{center}
\begin{minipage}{0.6\linewidth}
\begin{algorithm}[H]
    \renewcommand{\algorithmicrequire}{\textbf{Input:}}
    \renewcommand{\algorithmicensure}{\textbf{Output:}}
    \caption{\textbf{OFS-DPO Algorithm}}
    \label{algo1}
    \begin{algorithmic}[1]
        \REQUIRE SFT model $M_{SFT}$, Data stream $\mathcal{D}$, update $\bm{k}$
        \ENSURE Fast module param $\theta^{F}$
        \STATE Initialize F-module($M_{F}$), S-module($M_{S}$) with $M_{SFT}$
        \FOR {$i$ in $\bm{\mathcal{D}}$}
        \STATE $g_t^F = \nabla_{\theta^F} \mathcal L_{DPO-new}(\theta^F)$
        \STATE $g_t^S = \nabla_{\theta^S} \mathcal L_{DPO-new}(\theta^S)$
        \vspace{1pt}
        \STATE update $\theta^F,\theta^S$ with $g_t^F,g_t^S$ respectively
        \IF {$i\%\bm{k} == 0$}
        \IF {$\mathcal L_{DPO}(\theta^S)<\mathcal L_{DPO}(\theta^F)$}
        \STATE Interchange $M_{F}$, $M_{S}$
        \ENDIF
        \ENDIF
        \ENDFOR
    \end{algorithmic}
\end{algorithm}
\end{minipage}
\end{center}

\begin{algorithm}[H]
    \renewcommand{\algorithmicrequire}{\textbf{Input:}}
    \renewcommand{\algorithmicensure}{\textbf{Output:}}
    \caption{\textbf{COFS-DPO Algorithm}}
    \label{algo2}
    \setstretch{0.8}
    \textbf{STEP 1:}
    \vspace{2pt}
    \hrule
    \vspace{2pt}
    \begin{minipage}[t]{0.48\textwidth}
    \textbf{Task 1}
    \begin{algorithmic}[1]
        \REQUIRE $M_{SFT}$, Task1 data $\mathcal{D}_{1}$, update $\bm{k}$
        \ENSURE Fast module param $\theta_{1}^{F}$
        \STATE Initialize $M_{F}$, $M_{S}$ with $M_{SFT}$
        \vspace{1pt}
        \FOR {$i$ in $\mathcal{D}_{1}$}
        \STATE $g_t^F = \nabla_{\theta^F} \mathcal L_{DPO-new}(\theta^F)$
        \STATE $g_t^S = \nabla_{\theta^S} \mathcal L_{DPO-new}(\theta^S)$
        \vspace{1pt}
        \STATE update $M_{F},M_{S}$ with $g_t^F,g_t^S$ respectively
        \IF {$i\%\bm{k} == 0$}
        \IF {$\mathcal L_{DPO}(\theta^S)<\mathcal L_{DPO}(\theta^F)$}
        \STATE Interchange $M_{F}$, $M_{S}$
        \ENDIF
        \ENDIF
        \STATE Reserve data in $\mathcal M_1$ with randomness
        \ENDFOR
    \end{algorithmic}
    \end{minipage}
    \begin{minipage}[t]{0.48\textwidth}
    \textbf{Task 2}
    \begin{algorithmic}[1]
        \REQUIRE $\theta_{1}^{F}$, Task2 data $\mathcal{D}_{2}$, update $\bm{k}$
        \ENSURE Fast module param $\theta_{2}^{F}$
        \STATE Initialize $M_{F}$, $M_{S}$ with $\theta_{1}^{F}$
        \vspace{1pt}
        \FOR {$i$ in $\mathcal{D}_{2}$}
        \STATE $g_t^F = \nabla_{\theta^F} \mathcal L_{DPO-new}(\theta^F)$
        \STATE $g_t^S = \nabla_{\theta^S} \mathcal L_{DPO-new}(\theta^S)$
        \vspace{1pt}
        \STATE update $M_{F},M_{S}$ with $g_t^F,g_t^S$ respectively
        \IF {$i\%\bm{k} == 0$}
        \IF {$\mathcal L_{DPO}(\theta^S)<\mathcal L_{DPO}(\theta^F)$}
        \STATE Interchange $M_{F},M_{S}$
        \ENDIF
        \ENDIF
        \STATE Reserve data in $\mathcal M_2$ with randomness
        \ENDFOR
    \end{algorithmic}
    \end{minipage}
    \vspace{2pt}

    \hrule
    \vspace{2pt}
    \textbf{STEP 2:}
    \vspace{2pt}
    \hrule
    \vspace{2pt}
    \begin{algorithmic}
        \STATE $\beta^*_1\in(0,1)$, $\beta^*_2 \in (0,1)$, Using the retained data $\mathcal M_1, \mathcal M_2$, by COFS-DPO, look for $\beta^*_1, \beta^*_2$ that have the best performance of  generalization on the overall distribution after linear combination of the model parameter $\theta(\beta) = \beta^*_1\theta_{1}^{F}+\beta^*_2\theta_{2}^{F}$
    \end{algorithmic}
\end{algorithm}

\section{Experimental Details}
\label{Experimental Details}

\subsection{In-domain experiments}

In the in-domain task experiments, we employ distinct models and datasets across three tasks to assess the method's continual learning capability. Each experiment is limited to a single epoch, and all experiments are conducted on a single NVIDIA A800 80G GPU. The hyperparameters used in the experiments are detailed in Table \ref{in-domain-parameters}.
The evaluation methodology for GPT-4 remains consistent with DPO throughout the experiments. Each evaluation involves collecting 120 samples from the test set, with the prompts used during evaluation detailed in Table \ref{prompt-controlled}.

\begin{table}[H]
    \caption{Hyperparameters of different in-domain tasks.}
     \vspace{\baselineskip}
    \label{in-domain-parameters}
    \centering
    \resizebox{0.9\linewidth}{!}{
    \begin{tabular}{cccc}
    \toprule
    \textbf{Hyperparameters} & \textbf{Controlled sentiment generation} & \textbf{Summarization} & \textbf{Single-turn dialogue} \\
    \midrule
    model& gpt2-large & gptj & pythia28 \\
    batch size& 4 & 2 & 2 \\
    gradient accumulation steps& 1 & 2 & 2 \\
    seq length& 550 & 550 & 550 \\
    optimizer& adamw & adamw &  adamw \\
    slow lr& 5.00E-07 & 5.00E-07 & 5.00E-07 \\
    betas& [0.9, 0.999] & [0.9, 0.999] & [0.9, 0.999]  \\
    eps& 1.00E-08  & 1.00E-08 & 1.00E-08  \\
    weight decay& 1.00E-06 & 1.00E-06 & 1.00E-06  \\
    \bottomrule
    update $k$ & 10 & 5 & 10 \\
    loss weight $\alpha$& 0.1 & 0.7 & 0.7\\
    lr times & 2 & 2 & 2 \\
    \bottomrule
    \end{tabular}}
\end{table}

\begin{table}[H]
    \caption{Evaluation prompts for all in-domain tasks.}
     \vspace{\baselineskip}
    \label{prompt-controlled}
    \centering
    \resizebox{0.9\linewidth}{!}{
    \begin{tabularx}{\textwidth}{>{\hsize=0.2\hsize}XX}
        \toprule
        \makecell[c]{\textbf{Task}} & \makecell[c]{\textbf{Prompt}} \\
        \midrule
        \makecell[c]{Controlled \\[3pt] sentiment \\[3pt] generation} & \makecell[l]{
        Which of the following controlled sentiment generations does a better job of\\ generating the given text, without deviating from the text? A good generation \\
        is both positive and logical. \\ [4pt]
        prefixes: <test> \\ [4pt]
        generation A: <chosen> \\ [4pt]
        generation B: <model output> \\ [4pt]
        FIRST provide a one-sentence comparison of the two generations, explaining \\
        which you prefer and why. SECOND, on a new line, state only "A" or "B" to \\
        indicate your choice. Your response should use the format: \\
        Comparison: <one-sentence comparison and explanation> \\
        More positive: <"A" or "B">} \\
        \midrule
        \makecell[c]{Summarization} & \makecell[l]{
        Which of the following summaries does a better job of summarizing the most\\
        important points in the given forum post, without including unimportant or ir-\\
        relevant details? A good summary is both precise and concise. \\[4pt]
        Post: <test> \\[4pt]
        Summary A: <chosen> \\[4pt]
        Summary B: <model output> \\[4pt]
        FIRST provide a one-sentence comparison of the two summaries, explaining \\
        which you prefer and why. SECOND, on a new line, state only "A" or "B" to \\
        indicate your choice. Your response should use the format: \\
        Comparison: <one-sentence comparison and explanation> \\
        Preferred: <"A" or "B">} \\
        \midrule
        \makecell[c]{Single-turn \\[3pt] dialogue} & \makecell[l]{
        For the following query to a chatbot, which response is more helpful? \\[4pt]
        Query:  <user query> \\[4pt]
        Response A: <chosen> \\[4pt]
        Response B: <model output> \\[4pt]
        FIRST provide a one-sentence comparison of the two responses and explain  \\
        which is more helpful. SECOND, on a new line, state only "A" or "B" to in-\\
        dicate which response is more helpful. Your response should use the format: \\
        Comparison: <one-sentence comparison and explanation> \\
        More helpful: <"A" or "B">} \\
        \bottomrule
    \end{tabularx}}
\end{table}

\subsection{Cross-domain experiments}

In cross-domain experiments, we follow the task setting of CPPO, splitting the dataset into two task domains to assess the method's ability to retain old knowledge while learning new knowledge. In both COFS-DPO and the baseline method, experiments are conducted with two models: GPT2-s and LLaMA3. Once training is completed in both task domains, COFS-DPO aggregates the two LoRAs through weighted fusion to construct the final model. Experiments using GPT2-s can be completed on a single NVIDIA A800 80G GPU, whereas studies with LLaMA3 require two A800 GPUs.
The hyperparameters used in the experiments are shown in Table\ref{cross}. The metrics utilized for evaluation were adapted from the CPPO setup, with rPMs and ROUGE scores calculated based on the degree of alignment determined by the reference PM, as given in Table \ref{metrics}.

\begin{table}[H]
    \caption{The hyperparameters of various methods.}
    \label{cross}
     \vspace{\baselineskip}
    \centering
    \resizebox{0.6\linewidth}{!}{
    \begin{tabular}{cccc}
    \toprule
    \textbf{Hyperparameters} & \textbf{CPPOH} & \textbf{DPO} & \textbf{ours} \\
    \midrule
    model & \multicolumn{3}{c}{GPT2-s and Llama3} \\
    seq-length& 550 & 550 & 550 \\
    total steps& 25600 & - & - \\
    optimizer & adamw & adamw & adamw \\
    lr & 1.00E-05 & 5.00E-07 & 5.00E-07 \\
    betas & [0.9, 0.999] & [0.9, 0.999] & [0.9, 0.999] \\
    eps& 1.00E-08 & 1.00E-08 & 1.00E-08 \\
    weight-decay& 1.00E-06 & 1.00E-06 & 1.00E-06 \\
    \bottomrule
    update $k$ & - & - & 10 \\
    loss weight $\alpha$& - & - & 0.7\\
    lr times & - & - & 2 \\
    \bottomrule
    \end{tabular}}
\end{table}

\begin{table}[H]
    \caption{Metrics for our cross-domain tasks.}
    \label{metrics}
    \centering
     \vspace{\baselineskip}
    \resizebox{0.8\linewidth}{!}{
    \begin{tabular}{ccc}
    \toprule
    \textbf{Task ID} & \textbf{Metric} & \textbf{Definition} \\
    \midrule
    \multirow{2}{*}{\textbf{Task-1}} & rPM score on task-1($rPMS_1$) & $rPM(M_1, D_1^{test})$ \\
      & Rouge score on task-1($Rouge_1$) & $Rouge(M_1, D_1^{test})$ \\
    \midrule
    \multirow{5}{*}{\textbf{Final}} & rPM score on task-1($rPMS_1$) & $rPM(M_f, D_{1}^{test})$ \\
     & rPM score on task-2($rPMS_2$) & $rPM(M_f, D_2^{test})$ \\
     & Rouge score on task-1($Rouge_1$) & $Rouge(M_f, D_{1}^{test})$ \\
     & Rouge score on task-2($Rouge_2$) & $Rouge(M_f, D_{2}^{test})$ \\
     & Score Forgetting Ratio ($SFR$) & $rPM(M_1, D_1^{test}) - rPM(M_f, D_{1}^{test})$ \\
    \bottomrule
    \end{tabular}}
\end{table}

\end{document}